\documentclass{article}

\usepackage[numbers]{natbib}
\usepackage[final]{neurips_2024}

\usepackage[utf8]{inputenc} 
\usepackage[T1]{fontenc}    
\usepackage[pagebackref,breaklinks,colorlinks]{hyperref}
\usepackage{url}            
\usepackage{booktabs}       
\usepackage{amsfonts}       
\usepackage{nicefrac}       
\usepackage{microtype}      
\usepackage[dvipsnames]{xcolor}         
\usepackage{graphicx}
\usepackage{multirow}
\usepackage{wrapfig}
\usepackage{algpseudocode}
\usepackage{xspace}
\usepackage{float}
\usepackage{amsmath}
\usepackage{eccvabbrv}
\usepackage{caption}
\usepackage{enumitem}
\usepackage{subcaption}
\usepackage{algorithm}
\usepackage{algpseudocode}

\newcommand{\ours}{DiffuBox\xspace}
\newcommand{\ourslong}{\ours: Refining 3D Object Detection with Point Diffusion}

\newcommand{\lyft}{Lyft\xspace}
\newcommand{\ith}{Ithaca365\xspace}
\newcommand{\kitti}{KITTI\xspace}
\newcommand{\nuscenes}{nuScenes\xspace}

\newcommand{\na}{Direct\xspace}

\title{\ourslong{}}

\author{
\small{
Xiangyu Chen\thanks{Denotes equal contribution.} $^{,1,}$\thanks{Correspondences could be directed to \texttt{xc429@cornell.edu}}\hspace{8pt} 
Zhenzhen Liu\footnotemark[1] $^{,1}$\hspace{4pt}
Katie Z Luo\footnotemark[1] $^{,1}$\hspace{4pt}
Siddhartha Datta$^{2}$
} 
\\
\small{
\textbf{Adhitya Polavaram}$^{1}$\hspace{4pt}
\textbf{Yan Wang}$^{3}$\hspace{4pt}
\textbf{Yurong You}$^{3}$\hspace{4pt}
\textbf{Boyi Li}$^{3}$\hspace{4pt}
\textbf{Marco Pavone}$^{3}$
}
\\ 
\small{
\textbf{Wei-Lun Chao}$^{4}$\hspace{4pt}
\textbf{Mark Campbell}$^{1}$\hspace{4pt}
\textbf{Bharath Hariharan}$^{1}$\hspace{4pt}
\textbf{Kilian Q. Weinberger}$^{1}$}
\\
\small{
$^{1}$Cornell University\hspace{4pt}
$^{2}$University of Oxford\hspace{4pt}
$^{3}$NVIDIA Research\hspace{4pt}
$^{4}$The Ohio State University
}
}

\begin{document}

\maketitle

\begin{abstract}
Ensuring robust 3D object detection and localization is crucial for many applications in robotics and autonomous driving. 
%
Recent models, however, face difficulties in maintaining high performance when applied to domains with differing sensor setups or geographic locations, often resulting in poor localization accuracy due to domain shift. 
To overcome this challenge, we introduce a novel diffusion-based box refinement approach. 
This method employs a domain-agnostic diffusion model, conditioned on the LiDAR points surrounding a coarse bounding box, to simultaneously refine the box's location, size, and orientation. 
We evaluate this approach under various domain adaptation settings, and our results reveal significant improvements across different datasets, object classes and detectors.
Our PyTorch implementation is available at \href{https://github.com/cxy1997/DiffuBox}{https://github.com/cxy1997/DiffuBox}.
\end{abstract}
\section{Introduction}
\label{sec:intro}



3D object detection is a fundamental task for embodied agents to safely navigate in complex environments. 
For autonomous vehicles to navigate complicated traffic conditions, this amounts to identifying and localizing other road agents. 
Detection models under this setting need to make sense of LiDAR point clouds to identify accurate bounding boxes for pre-specified objects.
Given the diverse driving environments that occur in practice it is common, however, for the train- and test-time distributions to differ significantly. 
Domain distributional differences mainly arise from
differences in object size, point cloud density, and LiDAR beam angles. 
Consequently, models trained in one region or particular dataset (\eg Germany) may not perform well in another region or dataset (\eg USA)~\cite{wang2020train}. 
As a result, the domain adaptation problem raises concerns over the reliability and safety of 3D object detection in self-driving, that are often trained in a particular setting, then deployed into a diverse set of regions and locations.


Wang \etal~\cite{wang2020train}  
have obtained reductions in the domain adaptation gap by resizing boxes with a simple scaling heuristic after the fact. 
Consequently, we share the belief that the performance gap associated with domain adaptation is dominated by incorrect box sizes,  shapes, and orientations, rather than false positives and negatives in detections --- \eg a model trained in Germany can detect US cars, but struggles to capture their larger dimensions. 

In this paper we observe that, although the relationship of bounding boxes to the surrounding environment varies across domains, the relative position of LiDAR points with respect to their bounding boxes is surprisingly consistent~\cite{luo2024reward}.
Bounding boxes of these detections are, by definition, supposed to tightly fit the corresponding objects. 
Furthermore, the objects within the same object class (\eg, cars) have a similar shape with minor variance across different domains. 
What mostly varies, then, is the dimensions of the object, as opposed to this ``surface" shape when normalized to be the same size. 
Specifically, the distribution of points the LiDAR detector receives, is therefore consistent when normalized across object sizes, regardless of domains; points will always land near the edge of the bounding box no matter where the object is located. 
Thus, if we can somehow capture the distribution of points relative to a box's coordinate system, we would be able to use same process to fix incorrectly positioned bounding boxes to fit the correct point-distribution, even across domains.

    

Recognizing this observation, we propose \ours, a novel point diffusion model that learns the distribution of points relative to the object's bounding box in order to refine noisy bounding box proposals from the detection models for off-the-shelf domain adaptation. 
Given a set of noisy bounding box proposals, \ours denoises them into accurate detection boxes conditioned on the points near proposed bounding boxes.
Our method naturally avoids the domain gap caused by the scale difference~\cite{wang2020train}, since
\ours is designed to operate on object scale-invariant data, 
where we transform the LiDAR points around bounding box proposals into a normalized box view that is relative to the box instead of in absolute measure.
This eliminates the size priors presented in the source domain and forces the diffusion model to recover the accurate bounding box solely based on the relative position of points to the bounding box proposal, allowing for improved robustness in self-driving systems.

\begin{figure*}[t]
    \centering
    \includegraphics[width=\linewidth]{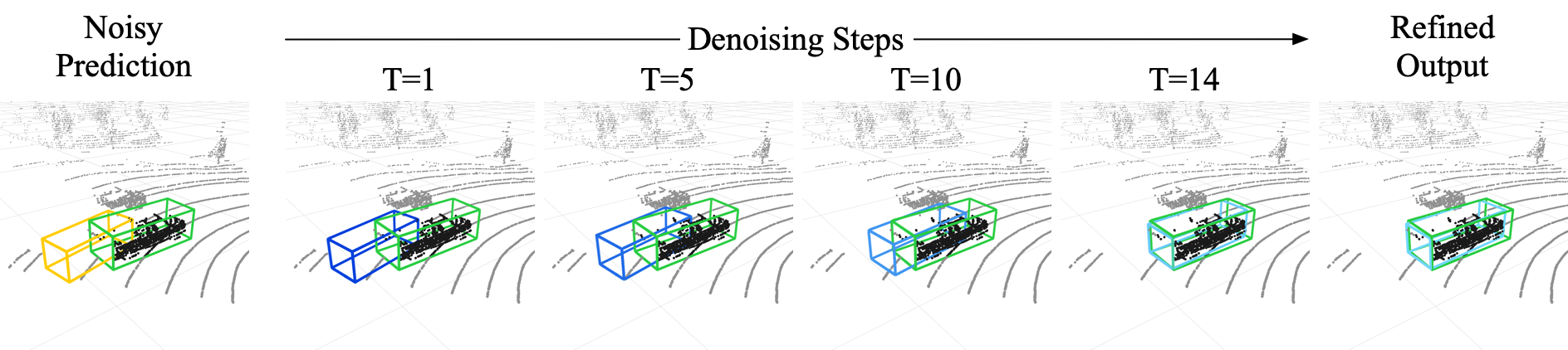}
    \caption{
    \textbf{Box refinement through denoising steps.} We visualize the correction of a noisy prediction, shown in yellow, using \ours. The ground truth box is visualized in green for reference. Boxes being refined are colored blue based on timestep. The output is refined iteratively though the denoising steps, resulting in the final, corrected output of our method.
    }
    \label{fig:qualitative}
    \vspace{-1.2\baselineskip}
\end{figure*}

To summarize, our contributions include:
We empirically validate our method, \ours, by adapting models trained on a dataset from Germany (KITTI \cite{kitti_dataset}) into two large, real-world datasets from the USA (Lyft L5 \cite{lyft_dataset} and Ithaca365 \cite{ithaca365}). 
Under both settings, we observe that \ours is able to refine the output bounding boxes drastically from the noisy initial predictions (\autoref{fig:qualitative}). 
Quantitatively, we observe strong improvements in mAP performance (up to 24 mAP), particularly in near-range boxes, where more points are present for \ours to refine the box predictions. 
When paired with a representative set of domain adaptation methods, including Output Transform, Statistical Normalization \cite{wang2020train}, and Rote-Domain Adaptation \cite{roteda}, \ours is able to further improve the results, and closing the gap between all method's final performance.

\section{Related Work}
\label{sec:related}

\paragraph{3D Object Detection.} In general, most 3D object detection methods require supervision from human-annotated data. 
They take 3D sensory data (e.g. LiDAR point clouds) and infer bounding boxes around 3D objects. 3D detection methods can be grouped into two categories based on the input representations: Point-based methods~\cite{qi2017pointnet, qi2017pointnet++, qi2018frustum, shi2019pointrcnn, shi2020point, yang20203dssd, pan20213d} that directly operate on point clouds, and grid-based methods~\cite{yan2018second, zhou2018voxelnet, lang2019pointpillars, shi2020pv, wang2020pillar, mao2021voxel} that first voxelize point clouds into 3D grids and then leverage convolutional architectures. 
Like other supervised models, 3D detection models suffer from decreased performance when the data distribution during inference differs from that during training. 
Our method \ours is designed to reduce the domain gap for general 3D object detection, agnostic to underlining model design.

\paragraph{Domain Adaptation in 3D.} Domain adaptation aims to alleviate the performance drop of 3D perception models under domain shift.
\cite{wang2020train} by Wang \etal is one of the first works studying the domain gap in 3D object detection and proposes Statistical Normalization (SN) that reduces the shape bias across domains.
ST3D~\cite{st3d}, Rote-DA~\cite{roteda}, and ST3D++~\cite{yang2022st3d++} propose a self-training pipeline that iteratively improve the target domain 3D detection performance with pseudo-label training and auxiliary priors.
Other methods can be grouped into feature-based~\cite{xu2021squeezesegv3,Langer2020DomainTF,morerio2017minimalentropy,kong2023conda, saltori2022cosmix,Kim_2023_CVPR} and architecture-based~\cite{jiang2021lidarnet,saito2018adversarial,vu2019advent,li2022domain, saleh2019domain} methods.
Some of them 
also apply data augmentation to construct and train domain-invariant representations to reduce the domain gap~\cite{kong2023conda, saltori2022cosmix,jiang2021lidarnet}.
Our proposed method is orthogonal to these methods and can be applied together with these models.

\paragraph{Diffusion Models.} 
%
Recently, diffusion models~\cite{sohl2015deep, ho2020denoising, song2020denoising, song2020score} have shown high-quality generative ability for image~\cite{dhariwal2021diffusion, latentdiffusion}, video~\cite{ho2022video, ho2022imagen} and 3D shape~\cite{luo2021diffusion, zeng2022lion, Melas-Kyriazi_2023_CVPR} modalities.
Zhou \etal~\cite{Zhou_2021_ICCV} uses diffusion models with a point-voxel representation for shape generation and point-cloud completion. 
LION \cite{zeng2022lion} uses a hierarchical VAE mapped to a latent space
and 
trains diffusion models on latent encodings to generate point clouds. 
In perception tasks, Chen \etal~\cite{chen2023diffusiondet} and Zhou \etal~\cite{zhou2023diffusion} propose diffusion-based object detection frameworks. Kim \etal~\cite{kim2023diffref3d} proposes a diffusion-based module to enhance the proposal refinement stage of two-stage object detectors. 
Unlike these approaches, our work focuses on leveraging diffusion for post-processing in a detector-agnostic manner that shows superior performance over previous methods.

\section{Method}
\label{sec:method}

\subsection{Problem Setup}
\label{sec:problem_setup}


Despite great in-domain performance, 3D object detection models often struggle to maintain their accuracy when generalized to new domains (datasets).
It has been concluded that such poor performance is mainly caused by mislocalization rather than misdetection~\cite{wang2020train}. 
That is, although objects can be correctly recognized by the object detector,  
the detected boxes 
lack sufficient overlap with the ground truth box and do not count as true positive (\ie, detections with IoU $< 0.7$ with ground-truth). 

In this work we introduce \ours, which focuses on correcting the localization of bounding box proposals, as illustrated in \autoref{fig:qualitative-full}, to improve domain adaptation for 3D object detection. 
Unlike existing domain adaptation algorithms that require careful re-training on the target~\cite{roteda, st3d} or source~\cite{wang2020train} domain data, \ours can be deployed off-the-shelf as a post-processing procedure in any novel domain.

Let $\boldsymbol{P} \in \mathcal{R}^{N \times 3}$ denote a $N$-point 3D point cloud from the target domain.
Let $\mathcal{B} = \{\boldsymbol{b}_1, \dots, \boldsymbol{b}_M\}$ be a set of $M$ imperfect bounding boxes proposed by an underadapted object detector given $\boldsymbol{P}$, where each bounding box $\boldsymbol{b}_i$ is a $7$-DoF (degrees of freedom) upright box, parameterized with center $\left[x_i, y_i, z_i\right]$, size $\left[w_i, l_i, h_i\right]$ and yaw angle $\theta_i$. 
We aim to obtain better localized object proposals $\hat{\mathcal{B}}$ by refining the boxes in $\mathcal{B}$ \emph{without any re-training}.
\begin{equation}
    \hat{\mathcal{B}} = \{\hat{\boldsymbol{b}}_1, \dots, \hat{\boldsymbol{b}}_M \mid  \hat{\boldsymbol{b}}_i = \text{refine}(\boldsymbol{b}_i, \boldsymbol{P})\}.
\end{equation}

\subsection{Learning Shapes in the Normalized Box View}
\label{sec:relevant_shape}

While domain differences between 3D object detection datasets exist in many aspects, the analysis from \cite{wang2020train} shows that the most significant hurdle for adaptation comes from the difference in object size. 
For instance, the American cars in the Lyft dataset~\cite{lyft_dataset} are about $20\%$ larger than German cars in the KITTI dataset~\cite{kitti_dataset} on average, and an object detector trained on KITTI will tend to still predict small boxes when tested on Lyft.
Unfortunately, as long as 3D object detectors are trained to explicitly predict object sizes, such size priors will be inevitably memorized during training and carried on to other domains as learned bias. 

\begin{figure*}[t]
    \centering
    \includegraphics[width=\linewidth]{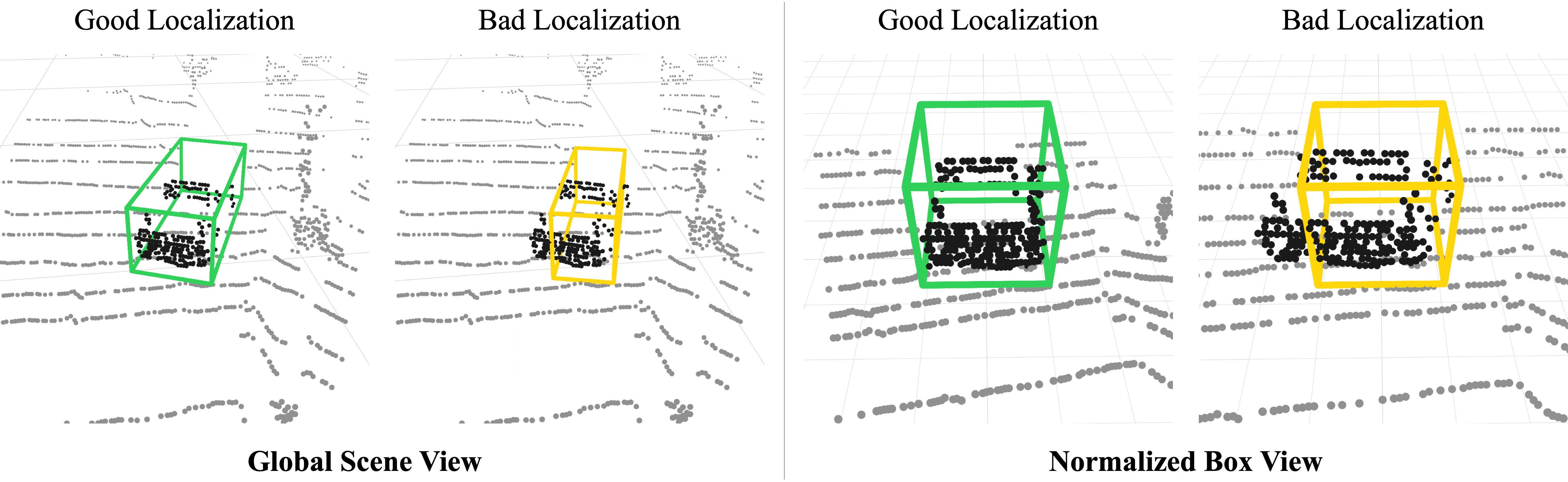}
    
  \caption{
  Example \emph{Car} objects converted into normalized box view (NBV). 
  Foreground/background points are marked in black/gray, respectively for better visualization. 
  Foreground LiDAR points distributing tightly within a ${\left[-1, 1\right]}^3$ NBV cube is a domain-consistent sign for good localization. }
  \label{fig:relevantshape}
  \vspace{-1.2\baselineskip}
\end{figure*}



We aim to achieve \textbf{scale-invariant object detection}, which would be naturally immune to size priors. 
Motivated by Luo~\etal{}'s ~\cite{luo2024reward} finding that the relative distribution of points to ground-truth bounding boxes is consistent across domains, \ie points tend to concentrate near the surface of boxes, we propose to disentangle object size from shape by transforming pointclouds into a normalized box view (NBV), where point coordinates are box-relative rather than absolute.


Using homogeneous transformation, we define $\boldsymbol{P}_{\boldsymbol{b}}^{\text{NBV}} \in \mathcal{R}^{N \times 3}$, the normalized box view of point cloud $\boldsymbol{P}$ relative to bounding box $\boldsymbol{b} = \left[x, y ,z, w, l, h, \theta\right]$, to be
\begin{equation}
\begin{bmatrix}
\boldsymbol{P}_{\boldsymbol{b}}^{\text{NBV}} \\
\boldsymbol{1}
\end{bmatrix} = 
\begin{bmatrix}
\frac{2}{l} & 0 & 0 & 0\\
0 & \frac{2}{w} & 0 & 0\\
0 & 0 & \frac{2}{h} & 0\\
0 & 0 & 0 & 1\\
\end{bmatrix}
\begin{bmatrix}
\cos\theta & \sin\theta & 0 & 0\\
-\sin\theta & \cos\theta & 0 & 0\\
0 & 0 & 1 & 0\\
0 & 0 & 0 & 1\\
\end{bmatrix}
\begin{bmatrix}
1 & 0 & 0 & -x\\
0 & 1 & 0 & -y\\
0 & 0 & 1 & -z\\
0 & 0 & 0 & 1\\
\end{bmatrix}
\begin{bmatrix}
\boldsymbol{P} \\
\boldsymbol{1}
\end{bmatrix}
\label{eq:nbv}
\end{equation}

As shown in \autoref{fig:relevantshape}, \autoref{eq:nbv} transforms the bounding box $\boldsymbol{b}$ into a ${\left[-1, 1\right]}^3$ cube, eliminating the size prior. 
The same transform also transforms $\boldsymbol{P}$ into box-relative, scale-invariant $\boldsymbol{P}_{\boldsymbol{b}}^{\text{NBV}}$. 

In practice, we only consider points within a certain depth range of the bounding box for efficiency. We refer to this range as \textit{context limit}. In the sections below, we overload $\boldsymbol{P}_{\boldsymbol{b}}^{\text{NBV}}$ as the point cloud within the context limit for ease of reference.



\subsection{Bounding Box Refinement via Diffusion}

Inspired by recent works on diffusion-based shape generation~\cite{yang2019pointflow, zhou20213d, luo2021diffusion, zeng2022lion} and knowledge distillation from pretrained diffusion models~\cite{poole2022dreamfusion}, we show that size-agnostic shape knowledge learned by a point cloud diffusion model can help to improve object localization across domains. The underlining assumption is that despite size difference, objects of the same category (\eg \emph{car}, \emph{cyclist}, \emph{pedestrian}) share similar shapes. 


\autoref{fig:relevantshape} illustrates our hypothesis that the good localization of a bounding box $\boldsymbol{b}$ is closely correlated to its corresponding $\boldsymbol{P}_{\boldsymbol{b}}^{\text{NBV}}$ forming a ``standard'' point distribution, \textit{a.k.a.} shape. 
Thus, improving the shape of $\boldsymbol{P}_{\boldsymbol{b}}^{\text{NBV}}$ will also lead to better localization of $\boldsymbol{b}$.
Since our ultimate goal is to optimize the bounding box $\boldsymbol{b} \rightarrow \hat{\boldsymbol{b}}$, we propose to use a point diffusion model to learn to ``denoise'' $\boldsymbol{P}_{\boldsymbol{b}}^{\text{NBV}} \rightarrow \boldsymbol{P}_{\hat{\boldsymbol{b}}}^{\text{NBV}}$.


Specifically, we begin by discussing the training of the diffusion model to learn the probabilistic flow for each point to a good box in Section \ref{sec:point_diffusion}.
Then, we go into how our method refines the bounding box by computing the improvement step relative to the learned probabilistic flow in Section \ref{sec:box_updates}.
Finally, we go into how we leverage the shape guidance to embed heuristics into our training procedure in Section \ref{sec:shape_guidance}. To enhance clarity, we additionally provide an algorithmic description of \ours's training and inference workflow in \autoref{sec:alg-desc}.

\subsubsection{Diffusion Training}
\label{sec:point_diffusion}

The learning objective of a diffusion model can be viewed as a variant of the score function~\cite{song2020score} $\nabla_{\boldsymbol{x}} \log{p\left(\boldsymbol{x}; \sigma\right)}$, where $\sigma$ indicates the noise level and $p\left(\boldsymbol{x}; 0\right) = p_{\text{data}}$, the true data distribution that is hard to directly sample from. 
As the score function points data to a higher likelihood, samples can instead be drawn from $p\left(\boldsymbol{x}; \sigma_{\text{max}}\right)$ --- which is usually modeled as an i.i.d. Gaussian distribution --- and denoised into $p_{\text{data}}$ by solving a probabilistic flow ODE~\cite{ince1956ordinary}/SDE~\cite{van1976stochastic}. 

Let $F_{\boldsymbol{\theta}}$ denote a diffusion model with parameter $\boldsymbol{\theta}$. 
Considering the full design space~\cite{karras2022elucidating}, in general its training loss can be written as:
\begin{equation}
    \mathbb{E}_{\sigma, \boldsymbol{y}, \boldsymbol{n}} 
    \left[
    \lambda(\sigma) c_{\text{out}}\left(\sigma\right)^2 
    \|F_\theta\left(
    c_{\text{in}}\left(\sigma\right) 
    \left(\boldsymbol{y} + \boldsymbol{n}\right);c_{\text{noise}}\left(\sigma\right)
    \right) - \frac{1}{c_{\text{out}}\left(\sigma\right)} \left(
    \boldsymbol{y} - c_{\text{skip}}\left(\sigma\right)
    \left(\boldsymbol{y} + \boldsymbol{n}\right)
    \right)\|_2^2\right]
    \text{,}
\end{equation}
where $\sigma \sim p_{\text{train}}$, $\boldsymbol{y}\sim p_{\text{data}}$, and $\boldsymbol{n}\sim\mathcal{N}\left(0, \sigma^2 \boldsymbol{I}\right)$. $\lambda(\sigma)$ denotes the effective training weight, $c_{\text{noise}}\left(\sigma\right)$ denotes noise level preconditioning. $c_{\text{in}}\left(\sigma\right)$, $c_{\text{out}}\left(\sigma\right)$, and $c_{\text{skip}}\left(\sigma\right)$ are input/output scaling factors. 

Modelling $p\left(\boldsymbol{x}; \sigma_{\text{max}}\right)$ as $\mathcal{N}\left(0, \boldsymbol{I}\right)$ allows for easy sampling.
However, an i.i.d. Gaussian noise doesn't suit the shift of point cloud in NBV caused by object mislocalization. 
As shown in \autoref{fig:by-step-normal}, a noisy NBV point cloud is formed from a 3D distortion on the standard shape, rather than adding Gaussian noise. 
The distortion includes rotation (caused by incorrect raw angle), rescaling (caused by incorrect size), and translation (caused by incorrect box center). 

Because of this, we made a few adaptations to the diffusion process.
We set the effective training weight $\lambda(\sigma) = 1$, all input/output scaling factors $c_{\text{in}}\left(\sigma\right) = c_{\text{out}}\left(\sigma\right) = c_{\text{skip}}\left(\sigma\right) = 1$, as the noise level $\sigma$ is unknown during inference. 
We apply Gaussian noise on the bounding box, rather than on the point cloud, to simulate mislocalized bounding boxes. 
With these adapatations, our new training loss becomes:
\begin{equation}
    \mathbb{E}_{\sigma, (\boldsymbol{P}, \boldsymbol{b^*}), \boldsymbol{n}} \left[\lambda(\sigma) \|F_\theta \left(\boldsymbol{P}_{\boldsymbol{b^*}+\boldsymbol{n}}^{\text{NBV}};c_{\text{noise}}(\sigma)
    \right) - \left(
    \boldsymbol{P}_{\boldsymbol{b^*}}^{\text{NBV}} - \boldsymbol{P}_{\boldsymbol{b^*}+\boldsymbol{n}}^{\text{NBV}}
    \right)\|_2^2\right]
\end{equation}
where $\sigma \sim p_{\text{train}}$, $\boldsymbol{P}, \boldsymbol{b^*} ~\sim \mathcal{D}_{\text{train}}$ and $\boldsymbol{n} \sim \mathcal{N}\left(0, \sigma^2 {diag}\left(\boldsymbol{\Sigma}\right)\right)$. 
$diag(\boldsymbol{\Sigma})$ is the variance of box noise, which is roughly estimated from \emph{direct} domain adaptation performance.





\subsubsection{Bounding Box Updates}
\label{sec:box_updates}
Since $F_{\boldsymbol{\theta}}$ is trained to approximate the score function $\nabla_{\boldsymbol{x}} \log{p\left(\boldsymbol{x}; \sigma\right)}$, the regular denoising process can be implemented by solving a probablistic flow ODE:
\begin{equation}
    \text{d}\boldsymbol{x} = -\dot{\sigma}(t) \sigma(t) \nabla_{\boldsymbol{x}} \log p \left(\boldsymbol{x}; \sigma(t)\right) \text{d} t\text{,}
\end{equation}
where $\sigma(t)$ denotes a noise schedule in which $\sigma(T) = \sigma_{\text{max}}$, $\sigma(0) = 0$, and the dot stands for a time derivative. 
Thus, $\boldsymbol{x}_0 \sim p_{\text{data}}$ can be generated by evolving $\boldsymbol{x}_T \sim p\left(\boldsymbol{x}; \sigma_{\text{max}}\right) = \mathcal{N}\left(0; \sigma^2\boldsymbol{I}\right)$ from time $t=T$ to $t=0$.

The NBV point cloud diffusion model $F_{\theta}\left( \boldsymbol{P}_{\boldsymbol{b}}^{\text{NBV}}; c_{\text{noise}}(\sigma)
    \right) \approx \frac{\nabla \log p\left(\boldsymbol{P}_{\boldsymbol{b}}^{\text{NBV}}; \sigma\right)}{\nabla {\boldsymbol{P}_{\boldsymbol{b}}^{\text{NBV}}}} $, to denoise the bounding box, rather than the point cloud, we take a further step following the chain rule ($\boldsymbol{P}_{\boldsymbol{b}}^{\text{NBV}}$ is differentiable according to Equation 2 of the main text):
\begin{equation}
    \frac{\log p\left(\boldsymbol{b}; \sigma, \boldsymbol{P}\right)}{\nabla {\boldsymbol{b}}} = \frac{\nabla \log p\left(\boldsymbol{P}_{\boldsymbol{b}}^{\text{NBV}}; \sigma\right)}{\nabla \boldsymbol{P}_{\boldsymbol{b}}^{\text{NBV}}} \frac{\nabla {\boldsymbol{P}_{\boldsymbol{b}}^{\text{NBV}}}}{\nabla {\boldsymbol{b}}} \approx F_{\theta}\left( \boldsymbol{P}_{\boldsymbol{b}}^{\text{NBV}}; c_{\text{noise}}(\sigma)
    \right) \frac{\nabla \boldsymbol{P}_{\boldsymbol{b}}^{\text{NBV}}}{\nabla {\boldsymbol{b}}}\text{.}
\end{equation}

Let $\boldsymbol{b}_T \sim p\left(\boldsymbol{b}; \sigma_{\text{max}}\right)$ be the imperfect bounding box predicted by an adapted object detector, and let $\boldsymbol{b}_0 \sim p\left(\boldsymbol{b}; 0\right)$ denote the corresponding box after refinement.
Similarly, bounding box refinement can be achieved by evloving $\boldsymbol{b}_T$ to $\boldsymbol{b}_0$ following:
\begin{equation}
    \text{d}\boldsymbol{b} = -\dot{\sigma}(t) \sigma(t) \frac{\nabla \log p\left(\boldsymbol{b}; \sigma, \boldsymbol{P}\right)}{\nabla \boldsymbol{b}} \text{d} t\text{,}
    \label{eq:box-ode}
\end{equation}

\subsubsection{Shape Guidance}
\label{sec:shape_guidance}
The probabilistic flow ODE allows adding objectives other than the score function to bounding box refinement without any retraining.
For instance, \cite{wang2020train} assumes the average object size ($\bar{w}, \bar{h}, \bar{l}$) in the target domain is available.
Such information can be used to further improve domain adaptation performance by simply rewriting \autoref{eq:box-ode} as:
\begin{equation}
    \text{d}\boldsymbol{b} = -\dot{\sigma}(t) \sigma(t) \left[\frac{\log p\left(\boldsymbol{b}; \sigma, \boldsymbol{P}\right)}{\nabla \boldsymbol{b}} +\alpha \frac{\nabla \ell_{\text{size}}\left(\boldsymbol{b}, \bar{w}, \bar{h}, \bar{l} \right)}{\nabla \boldsymbol{b}}\right] \text{d} t\text{,}
\end{equation}
where $\alpha$ denotes shape weight, and
\begin{equation}
    \ell_{\text{size}}\left(\boldsymbol{b}, \bar{w}, \bar{h}, \bar{l} \right) = {\|w - \bar{w}\|}^2 + {\|h - \bar{h}\|}^2 + {\|l - \bar{l}\|}^2\text{.}
\end{equation}

\section{Experiments}
\subsection{Experimental Setup}
\paragraph{Datasets.} We primarily consider three datasets: The \kitti dataset~\cite{kitti}, the \lyft Level 5 Perception dataset~\cite{lyft}, and the \ith dataset~\cite{ithaca365}. For \kitti, we follow the official splits. For \lyft, we follow various existing works~\cite{modest, roteda, drift} and use the splits separated by geographical locations, consisting of 11,873 point clouds for training and 4,901 for testing. For \ith, we utilize the annotated point clouds with 4,445 for training and 1,644 for testing. Additionally, we include experiments with the \nuscenes dataset in the supplementary to evaluate \ours's performance on larger-scale and more diverse data. 

\paragraph{Baselines.} We consider five domain adaptation baselines: (1) directly applying an out-of-domain detector without adaptation (\na); (2) Output Transformation (OT)~\cite{wang2020train}; (3) Statistical Normalization (SN)~\cite{wang2020train}; (4) Rote-DA~\cite{roteda}; (5) ST3D~\cite{st3d}. OT and SN perform resizing based on the average sizes from the target domain. OT directly resizes the predicted bounding boxes on the target domain, while SN trains the detector with resized objects and boxes from the source domain. Rote-DA and ST3D perform self-training. Rote-DA leverages an additional context in the form of persistency-prior~\cite{barnes2018driven} and enforces consistency across domains. ST3D leverages better data augmentation and a memory bank for high-quality detections. As \ours is complementary to these methods, we compare the detection performance of these methods before and after refining with \ours. 


\paragraph{Evaluation Metrics.} We evaluate the detection performance in Bird's Eye View (BEV) and 3D. At depth ranges of 0-30m, 30-50m and 50-80m, we report the mean Average Precision (mAP) with Intersection over Union (IoU) thresholds set at 0.7 for cars, and 0.5 for pedestrians and cyclists. We also consider the nuScenes true positive metrics \cite{nuscenes}: translation error, scale error and rotation error. These measure the error in center offset, size difference, and orientation offset, respectively, of all true positive detections. 

\paragraph{Implementation Details.} We use the implementation and configurations from OpenPCDet~\cite{openpcdet2020} for detectors, and \cite{edm}'s implementation for diffusion models. We set the context limit to 4x the bounding box size. We use shape weight 0.1 for cars and pedestrians, and 0.01 for cyclists as cyclists have more shape variation. More details can be found in the supplementary. 

\subsection{Experimental Results}
We present the results for \kitti $\rightarrow$ \lyft cars in \autoref{tab:adaptation} (mAP@IoU 0.7) and \autoref{tab:nuscenes_tp} (\nuscenes TP metrics), and the results for \kitti $\rightarrow$ \ith cars in \autoref{tab:adaptation_ith} (mAP@IoU 0.7) and \autoref{tab:nuscenes_tp_ith} (\nuscenes TP metrics). We additionally include \kitti $\rightarrow$ \nuscenes results in \autoref{tab:adaptation_car_nusc} in the supplementary. We use PointRCNN~\citep{shi2019pointrcnn} detectors; evaluations with other detectors can be found in the section below. \ours consistently attains significant performance gain across different domain adaptation methods and datasets. The improvement of \ours is especially significant for near-range and middle-range detections. Notably, for \kitti $\rightarrow$ \lyft cars, \ours applied upon the \textit{\na} outputs is able to attain comparable performance with domain adaptation methods that require training such as ST3D. We hypothesize that this is because there are more LiDAR points for near-range and middle-range objects, which allows \ours to better correct the detections.  

\begin{table*}[!t]
\captionsetup{aboveskip=0.5em}
\centering
\caption{\textbf{mAP@IoU 0.7 for \kitti $\rightarrow$ \lyft (cars).} Higher is Better. \ours leads to improvement in almost all cases, with especially significant gain for the Direct and OT detections.}
  \resizebox{\linewidth}{!}
{

\begin{tabular}{l|cccc|cccc}
\toprule
\multirow{2}{*}{Method} & \multicolumn{4}{c}{BEV$\uparrow$} & \multicolumn{4}{c}{3D$\uparrow$} \\
\cmidrule{2-9}
& \ 0-30m \  & \ 30-50m\  & \ 50-80m\  & \ 0-80m\  & \ 0-30m\  & \ 30-50m\  & \ 50-80m\  & \ 0-80m\  \\
\midrule
Direct & 68.03 & 38.62 & 9.99 & 39.06 & 25.76 & 7.84 & 1.04 & 12.07 \\
Direct+\ours & \textbf{88.95} & \textbf{73.27} & \textbf{23.84} & \textbf{59.70} & \textbf{62.94} & \textbf{35.44} & \textbf{6.67} & \textbf{35.56} \\\midrule
OT & 75.07 & 61.84 & 20.44 & 51.95 & 18.67 & 10.57 & 1.82 & 11.89 \\
OT+\ours & \textbf{92.67} & \textbf{74.46} & \textbf{23.95} & \textbf{60.98} & \textbf{50.99} & \textbf{33.06} & \textbf{6.87} & \textbf{31.21} \\\midrule
SN & 92.88 & 69.97 & \textbf{25.68} & 61.67 & \textbf{70.40} & 32.96 & 6.18 & 36.64 \\
SN+\ours & \textbf{94.77} & \textbf{72.09} & 25.47 & \textbf{62.70} & 69.62 & \textbf{40.39} & \textbf{7.17} & \textbf{38.72}\\ \midrule
Rote-DA & 89.64 & 70.10 & \textbf{27.96} & 60.63 & 50.65 & 24.92 & \textbf{7.43} & 28.63\\
Rote-DA+\ours & \textbf{95.10} & \textbf{75.10} & 23.35 & \textbf{62.14} & \textbf{71.00} & \textbf{48.89} & 6.48 & \textbf{41.06}\\ \midrule
ST3D & 72.86 & 64.23 & 34.96 & 55.58 & 35.22 & 26.33 & 6.06 & 22.47 \\
ST3D+\ours & \textbf{92.08} & \textbf{75.03} & \textbf{35.35} & \textbf{66.08} & \textbf{61.58} & \textbf{45.44} & \textbf{10.16} & \textbf{39.81} \\
\bottomrule
\end{tabular}
}
\label{tab:adaptation}
\end{table*} %

\begin{table*}[!t]
\captionsetup{aboveskip=0.5em}
\centering
\caption{\textbf{mAP@IoU 0.7 for \kitti $\rightarrow$ \ith (cars).} \ours leads to significant improvement upon different adaptation methods. }
  \resizebox{\linewidth}{!}
{

\begin{tabular}{l|cccc|cccc}
\toprule
\multirow{2}{*}{Method} & \multicolumn{4}{c}{BEV$\uparrow$} & \multicolumn{4}{c}{3D$\uparrow$} \\
\cmidrule{2-9}
& \ 0-30m\  & \ 30-50m\  & \ 50-80m\  & \ 0-80m\  & \ 0-30m\  & \ 30-50m\  & \ 50-80m\  & \ 0-80m\  \\
\midrule
Direct & 52.59 & 21.19 & 3.20 & 25.08 & 25.09 & 6.25 & 0.17 & 10.53 \\
Direct+\ours & \textbf{61.89} & \textbf{32.09} & \textbf{6.05} & \textbf{32.27} & \textbf{42.23} & \textbf{17.79} & \textbf{1.47} & \textbf{20.51} \\\midrule
OT & 59.34 & 29.18 & 5.26 & 30.11 & 32.05 & 12.00 & 1.16 & 14.71 \\
OT+\ours & \textbf{60.76} & \textbf{32.56} & \textbf{6.07} & \textbf{31.89} & \textbf{40.43} & \textbf{18.33} & \textbf{1.61} & \textbf{19.80}\\\midrule
SN & 60.48 & 31.04 & \textbf{4.04} & 29.80 & 32.17 & 13.03 & 0.85 & 15.02 \\
SN+\ours & \textbf{60.79} & \textbf{34.49} & 3.79 & \textbf{30.81} & \textbf{37.21} & \textbf{18.90} & \textbf{1.33} & \textbf{18.31} \\\midrule
Rote-DA & 71.14 & 44.76 & 14.00 & \textbf{42.38} & 43.07 & 22.42 & 2.46 & 22.38\\
Rote-DA+\ours & \textbf{71.52} & \textbf{45.44} & \textbf{14.56} & 42.28 & \textbf{46.77} & \textbf{25.72} & \textbf{4.17} & \textbf{25.02}\\
\bottomrule
\end{tabular}
}
\vspace{-0.5\baselineskip}
\label{tab:adaptation_ith}
\end{table*} %

\paragraph{\ours with Other Detectors.}
To further demonstrate the robustness and versatility of \ours, we present \ours's performance for refining predictions from other detectors. We consider PointPillar~\cite{lang2019pointpillars}, SECOND~\cite{yan2018second}, PV-RCNN~\cite{shi2020pv}, CenterPoint~\cite{yin2021center} and DSVT~\cite{wang2023dsvt} trained on \kitti. We report the mAP@IoU 0.7 for \kitti $\rightarrow$ \lyft cars in \autoref{tab:adaptation_other_detectors}. Results show that \ours consistently improves the predictions from different detectors. 
\begin{table*}[!t]
\captionsetup{aboveskip=0.5em}
\centering
\caption{\textbf{mAP@IoU 0.7 for \kitti $\rightarrow$ \lyft (cars) with other detectors.} \ours consistently improves the detections from different detectors. }
  \resizebox{\linewidth}{!}
{

\begin{tabular}{l|cccc|cccc}
\toprule
\multirow{2}{*}{Method} & \multicolumn{4}{c}{BEV$\uparrow$} & \multicolumn{4}{c}{3D$\uparrow$} \\
\cmidrule{2-9}
& \ 0-30m \  & \ 30-50m\  & \ 50-80m\  & \ 0-80m\  & \ 0-30m\  & \ 30-50m\  & \ 50-80m\  & \ 0-80m\  \\
\midrule
PointPillar (Direct) & 65.77 & 39.02 & 11.28 & 36.80 & 16.58 & 5.06 & 0.56 & 6.87 \\
PointPillar (Direct)+\ours & \textbf{84.76} & \textbf{65.57} & \textbf{17.46} & \textbf{53.67} & \textbf{65.41} & \textbf{32.77} & \textbf{4.55} & \textbf{33.82} \\\midrule
SECOND (Direct) & 65.61 & 38.83 & 13.92 & 38.06 & 24.39 & 8.32 & 0.86 & 10.68\\
SECOND (Direct)+\ours & \textbf{89.70} & \textbf{66.98} & \textbf{19.11} & \textbf{57.04} & \textbf{61.90} & \textbf{32.24} & \textbf{4.39} & \textbf{33.19}\\\midrule
PV-RCNN (Direct)& 73.56 & 46.39 & 13.63 & 43.72 & 34.20 & 14.03 & 1.53 & 16.17 \\
PV-RCNN (Direct)+\ours & \textbf{92.40} & \textbf{68.35} & \textbf{20.68} & \textbf{59.30} & \textbf{63.82} & \textbf{35.78} & \textbf{5.03} & \textbf{35.43} \\\midrule
PV-RCNN (OT) & 80.48 & 54.40 & 17.54 & 51.03 & 19.87 & 8.15 & 0.75 & 10.74 \\
PV-RCNN (OT)+\ours & \textbf{93.65} & \textbf{69.06} & \textbf{21.35} & \textbf{60.16} & \textbf{53.68} & \textbf{31.31} & \textbf{4.00} & \textbf{30.63} \\\midrule
PV-RCNN (SN) & \textbf{94.16} & 68.58 & \textbf{22.22} & \textbf{62.16} & \textbf{72.72} & 27.86 & 3.43 & 33.47\\
PV-RCNN (SN)+\ours & 93.99 & \textbf{69.24} & 21.13 & 61.23 & 67.83 & \textbf{35.08} & \textbf{4.50} & \textbf{34.93}\\\midrule
CenterPoint (Direct) & 74.91 & 36.64 & 2.47 & 36.23 & 28.63 & 4.05 & 0.15 & 10.29\\
CenterPoint (Direct)+DiffuBox & \textbf{90.25} & \textbf{58.65} & \textbf{6.95} & \textbf{51.38} & \textbf{71.02} & \textbf{34.91} & \textbf{1.48} & \textbf{34.40}\\\midrule
CenterPoint (OT) & 82.81 & 51.02 & 6.26 & 46.10 & 25.34 & 12.35 & 0.52 & 13.34\\
CenterPoint (OT)+DiffuBox & \textbf{91.79} & \textbf{59.24} & \textbf{7.45} & \textbf{51.88} & \textbf{63.01} & \textbf{32.53} & \textbf{1.24} & \textbf{30.95}\\\midrule
DSVT (Direct) & 68.93 & 47.49 & 11.32 & 41.77 & 33.72 & 11.74 & 1.42 & 15.67\\
DSVT (Direct)+DiffuBox & \textbf{89.01} & \textbf{63.41} & \textbf{17.50} & \textbf{55.27} & \textbf{65.22} & \textbf{36.31} & \textbf{5.02} & \textbf{35.61}\\\midrule
DSVT (OT) & 71.85 & 42.93 & 13.18 & 43.05 & 15.66 & 4.47 & 0.31 & 8.06\\
DSVT (OT)+DiffuBox & \textbf{90.21} & \textbf{63.50} & \textbf{17.91} & \textbf{55.61} & \textbf{56.84} & \textbf{31.19} & \textbf{4.63} & \textbf{31.12}\\
\bottomrule
\end{tabular}
}
\vspace{-1.2\baselineskip}
\label{tab:adaptation_other_detectors}
\end{table*} %

\paragraph{\ours on Other Object Classes.}
We present \ours's performance on other object classes, specifically pedestrians and cyclists. We use the same configurations as cars, except cyclist shape weight 0.01 as mentioned in the implementation details. We report the \kitti $\rightarrow$ \lyft performance in \autoref{tab:adaptation_pedestrian_cyclist}, and the \kitti $\rightarrow$ \ith performance in \autoref{tab:adaptation_pedestrian_ith}. For \ith, we only evaluate for pedestrians as \ith has very few cyclists. Results show that \ours consistently attains significant improvement across object classes and domains, and can improve upon other domain adaptation methods. This shows the robustness and versatility of \ours.

\begin{table*}[!th]
\captionsetup{aboveskip=0.5em}
\centering
\caption{\textbf{mAP@IoU 0.5 for \kitti $\rightarrow$ \lyft (pedestrians/cyclists).} \ours attains consistent improvement across object classes and is able to improve upon other domain adaptation methods. }
  \resizebox{\linewidth}{!}
{

\begin{tabular}{l|l|cccc|cccc}
\toprule
\multirow{2}{*}{Class} & \multirow{2}{*}{Method} & \multicolumn{4}{c}{BEV$\uparrow$} & \multicolumn{4}{c}{3D$\uparrow$} \\
\cmidrule{3-10}
& & \ 0-30m\  & \ 30-50m\  & \ 50-80m\  & \ 0-80m\  & \ 0-30m\  & \ 30-50m\  & \ 50-80m\  & \ 0-80m\  \\
\midrule
\multirow{8}{*}{ Pedestrian }  & Direct & 31.89 & 25.75 & 0.51 & 20.74 & 21.29 & 16.59 & 0.18 & 14.17 \\
 & Direct+\ours & \textbf{43.64} & \textbf{26.97} & \textbf{0.61} & \textbf{25.10} & \textbf{34.30} & \textbf{22.54} & \textbf{0.33} & \textbf{20.46} \\\cmidrule{2-10}
 & OT & 35.63 & \textbf{25.47} & \textbf{0.64} & 21.69 & 27.76 & 19.02 & 0.35 & 16.77 \\
& OT+\ours &  \textbf{42.63} & 25.28 & 0.55 & \textbf{24.02} & \textbf{37.34} & \textbf{22.16} & \textbf{0.42} & \textbf{20.94} \\\cmidrule{2-10}
& SN & 43.75 & \textbf{36.87} & 0.67 & 28.18 & 34.12 & 26.08 & 0.45 & 21.00\\
& SN+\ours & \textbf{50.98} & 35.88 & \textbf{1.08} & \textbf{29.45} & \textbf{38.55} & \textbf{30.24} & \textbf{0.60} & \textbf{23.50} \\\cmidrule{2-10}
&Rote-DA & 49.14 & 46.86 & \textbf{1.23} & 33.60 & 37.68 & 39.25 & 1.01 & 26.75 \\
&Rote-DA+\ours & \textbf{54.14} & \textbf{50.30} & 1.22 & \textbf{36.04} & \textbf{42.76} & \textbf{43.25} & \textbf{1.04} & \textbf{29.93} \\\midrule
\multirow{8}{*}{ Cyclist }  & Direct & 48.50 & 8.91 & \textbf{0.13} & 26.96 & 38.13 & 5.19 & \textbf{0.02} & 21.11 \\
 & Direct+\ours & \textbf{61.28} & \textbf{10.60} & 0.06 & \textbf{34.90} & \textbf{49.76} & \textbf{6.87} & \textbf{0.02} & \textbf{27.35} \\\cmidrule{2-10}
 & OT & 55.37 & 9.99 & \textbf{0.14} & 30.90 & 20.09 & 4.11 & \textbf{0.02} & 10.96\\
& OT+\ours & \textbf{65.93} & \textbf{10.26} & 0.06 & \textbf{36.86} & \textbf{32.93} & \textbf{6.11} & \textbf{0.02} & \textbf{18.81}\\\cmidrule{2-10}
& SN & 46.75 & 11.42 & 0.05 & 26.30 & 36.87 & 6.23 & \textbf{0.03} & 20.18\\
& SN+\ours & \textbf{59.48} & \textbf{15.82} & \textbf{0.10} & \textbf{34.62} & \textbf{48.92} & \textbf{10.31} & 0.02 & \textbf{27.38} \\\cmidrule{2-10}
&Rote-DA & 77.19 & \textbf{34.61} & \textbf{0.09} & 48.66 & 70.35 & \textbf{30.77} & \textbf{0.05} & 44.29\\
&Rote-DA+\ours & \textbf{83.80} & 31.36 & 0.05 & \textbf{51.49} & \textbf{75.79} & 27.98 & 0.04 & \textbf{45.77} \\
\bottomrule
\end{tabular}
}
\vspace{-1.2\baselineskip}
\label{tab:adaptation_pedestrian_cyclist}
\end{table*} %




\subsection{Qualitative Results}
\begin{figure*}[t]
\centering
\includegraphics[width=\linewidth]{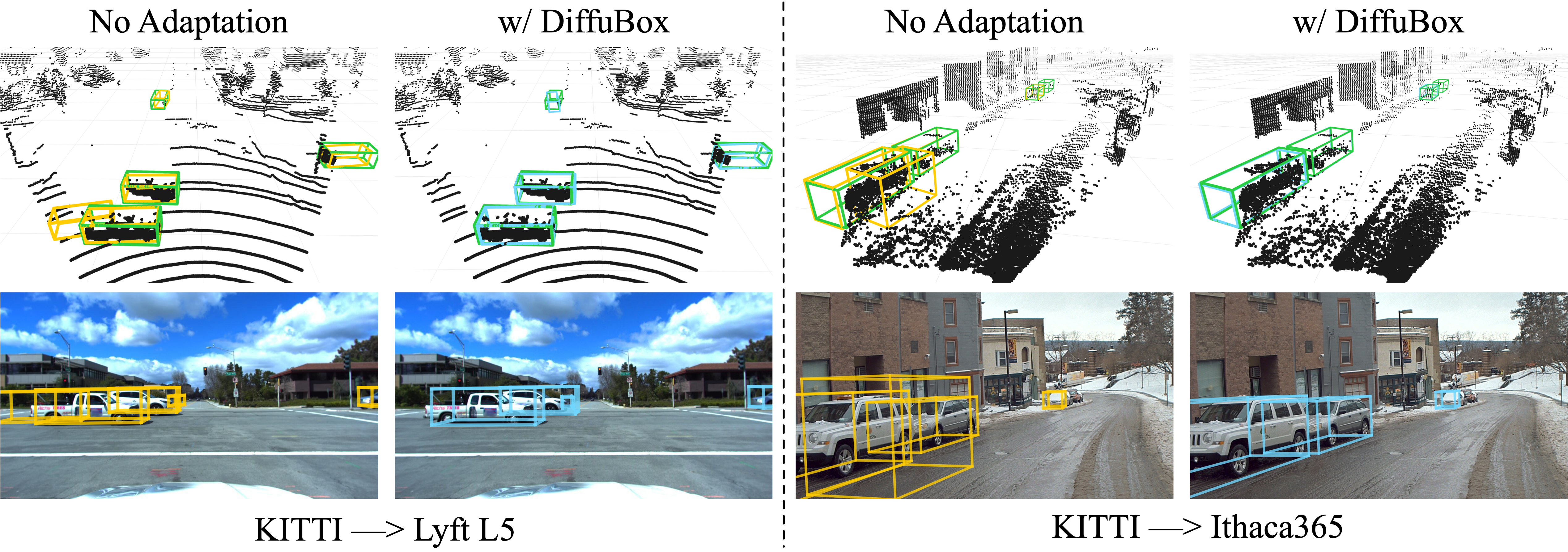}
\caption{
\textbf{Illustration of 3D object detection on \lyft/\ith before and after \ours's refinement.}
We visualize detections from an out-of-domain PointRCNN on four scenes from each dataset. 
We color the ground truth boxes in green, the detector outputs in yellow, and \ours's refinements in blue. 
The out-of-domain detector sometimes produces false positives or boxes with incorrect shape or alignment. 
\ours effectively improves the wrong or inaccurate boxes, while making little change to the accurate boxes. }
\label{fig:qualitative-full}
\vspace{-1.2\baselineskip}
\end{figure*}

\autoref{fig:qualitative-full} visualizes four scenes from the \lyft and \ith datasets. We compare the ground truth bounding boxes (green), the detections directly obtained from a PointRCNN trained on \kitti (yellow), and the refined detections using \ours (blue). The out-of-domain PointRCNN produces reasonable results, but occasionally it produces false positives or boxes with incorrect shapes or alignment. \ours effectively moves the incorrect boxes towards having better location, shape and alignment. Also observe that for the already accurate boxes, \ours makes little change to them. 

\subsection{\ours Extension: Detector Retraining}

\begin{wrapfigure}{r}{0.5\textwidth}
\vspace{-2\baselineskip}
  \begin{center}
    \includegraphics[width=0.5\textwidth]{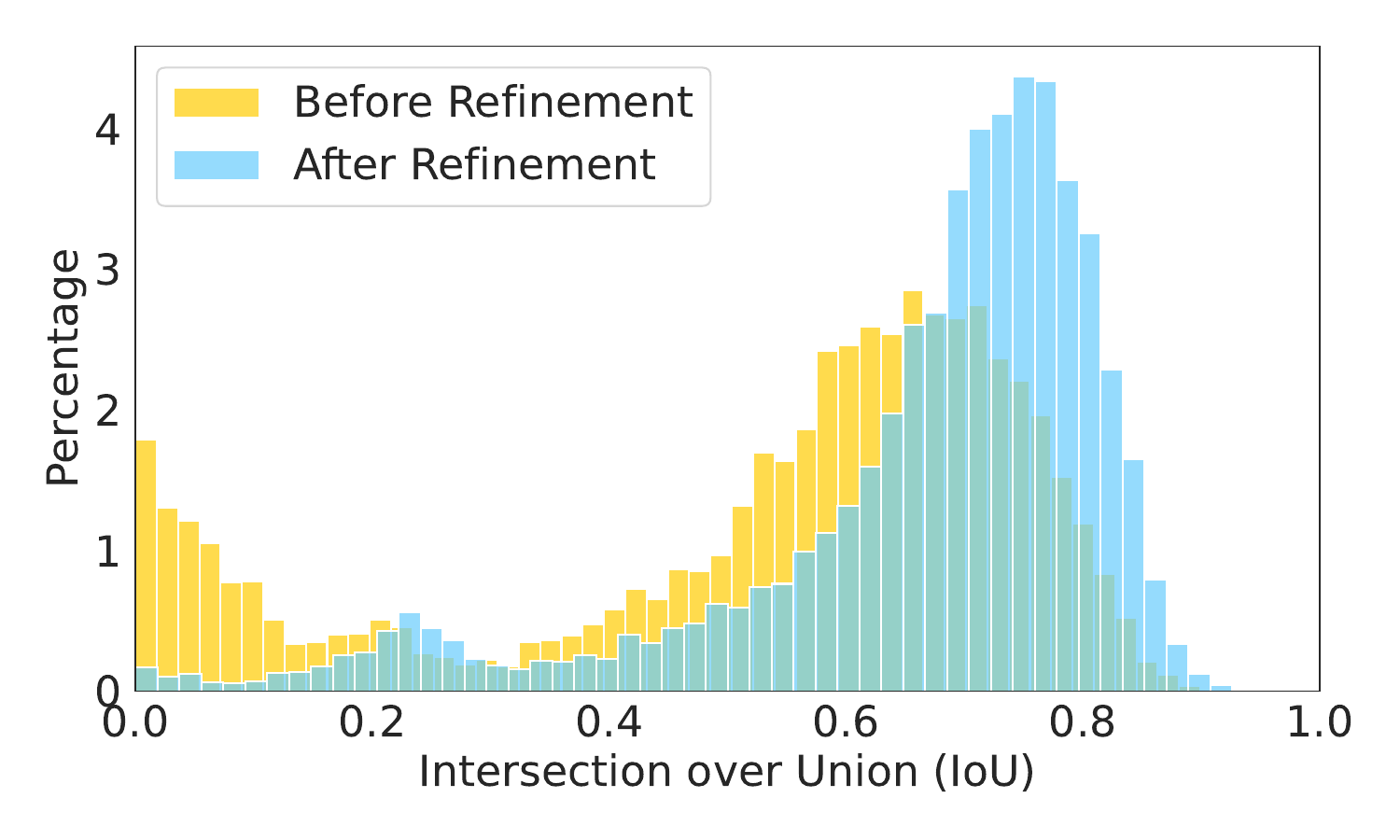}
  \end{center}
    \vspace{-\baselineskip}
  \caption{\textbf{Comparison of bounding box quality before and after refinement with \ours.} We report the distribution of Intersection over Union (IoU) with ground-truth labels from the \lyft dataset. The unrefined predictions are from an unadapted Point-RCNN model trained on \kitti. We show that \ours leads to significant improvement in bounding box localization.}
  \label{fig:iou-comp}
  \vspace{-2\baselineskip}
\end{wrapfigure}

One extension of \ours for domain adaptation is to retrain detectors~\cite{you2022exploiting, modest} with \ours's refined boxes. We can take a detector trained on the source domain, obtain its predictions on the target domain, refine the predictions with \ours, and then retrain a detector using the refined boxes as labels. 

We provide the \kitti $\rightarrow$ \lyft results in \autoref{tab:retraining}. We compare the performance of (1) directly applying the \kitti detector, (2) retraining for one round with the \kitti detector's predictions as labels, and (3) retraining for one round with the \kitti detector's predictions after \ours's refinement as labels. Retraining is performed based on the \kitti detector's predictions on the \lyft training split, and the evaluation is conducted over \lyft's testing split. Retraining using the \kitti detector's predictions directly only provides limited improvement over directly applying the \kitti detector, while retraining with \ours's refinement leads to significant improvement. 

\begin{table*}[!ht]
\captionsetup{aboveskip=0.5em}
\centering
\caption{\textbf{Retraining Performance (mAP@IoU 0.7) for \kitti $\rightarrow$ \lyft.} Retraining with \ours's refined detections attains significant performance improvement. }
  \resizebox{\linewidth}{!}
{
\begin{tabular}{l|l|cccc|cccc}
\toprule
\multirow{2}{*}{Class} & \multirow{2}{*}{Method} & \multicolumn{4}{c}{BEV$\uparrow$} & \multicolumn{4}{c}{3D$\uparrow$} \\
\cmidrule{3-10}
& & \ 0-30m \  & \ 30-50m\  & \ 50-80m\  & \ 0-80m\  & \ 0-30m\  & \ 30-50m\  & \ 50-80m\  & \ 0-80m\  \\\midrule
\multirow{3}{*}{ Car}& Direct & 68.03 & 38.62 & 9.99 & 39.06 & 25.76 & 7.84 & 1.04 & 12.07 \\
& Retraining w/ Direct &  70.98 & 45.53 & 15.12 & 43.49 & 30.45 & 10.54 & 1.69 & 14.55\\
& Retraining w/ Direct+\ours & \textbf{91.79} & \textbf{77.52} & \textbf{35.33} & \textbf{64.71} & \textbf{67.80} & \textbf{36.17} & \textbf{10.56} & \textbf{37.55} \\\midrule
\multirow{3}{*}{ Pedestrian}& Direct & 31.89 & 25.75 & 0.51 & 20.74 & 21.29 & 16.59 & 0.18 & 14.17 \\
& Retraining w/ Direct & 34.23 & 34.65 & \textbf{0.95} & 24.37 & 27.58 & 30.33 & \textbf{0.37} & 20.22 \\
& Retraining w/ Direct+\ours & \textbf{44.17} & \textbf{39.39} & 0.84 & \textbf{29.22} & \textbf{35.60} & \textbf{34.67} & 0.36 & \textbf{24.71}\\  \midrule
\multirow{3}{*}{ Cyclist}& Direct & 48.50 & \textbf{8.91} & 0.13 & 26.96 & 38.13 & 5.19 & 0.02 & 21.11 \\
& Retraining w/ Direct &  44.06 & 6.05 & 0.21 & 24.29 & 33.55 & \textbf{5.73} & \textbf{0.14} & 18.73\\
& Retraining w/ Direct+\ours & \textbf{57.57} & 6.54 & \textbf{0.23 }& \textbf{31.36} & \textbf{45.79} & 5.19 & 0.11 & \textbf{24.39} \\ 
\bottomrule
\end{tabular}
}
\label{tab:retraining}
\end{table*} %

\subsection{Ablation Studies and Analysis}
\paragraph{Context Limit.} We conduct ablation study on the context limit, and consider ranges of 2x, 4x and 6x the box size. We evaluate under the setting of no adaptation (\textit{\na}) for \kitti $\rightarrow$ \lyft  cars, and compare the mAP@IoU 0.7 in~\autoref{tab:context_limit_abl}. All three context limits lead to significant performance improvement. Larger limit attains more gain, and the gain saturates at around the 4x limit. 

\begin{table*}[!t]
\captionsetup{aboveskip=0.5em}
\centering
\caption{\textbf{Ablation on Context Limit.} \ours is robust against the choice of context limit. Larger limit could lead to better performance, with the gain saturated at around 4x limit. }
  \resizebox{\linewidth}{!}
{

\begin{tabular}{l|cccc|cccc}
\toprule
\multirow{2}{*}{Method} & \multicolumn{4}{c}{BEV$\uparrow$} & \multicolumn{4}{c}{3D$\uparrow$} \\
\cmidrule{2-9}
& \ 0-30m\  & \ 30-50m\  & \ 50-80m\  & \ 0-80m\  & \ 0-30m\  & \ 30-50m\  & \ 50-80m\  & \ 0-80m\  \\
\midrule
Direct & 68.03 & 38.62 & 9.99 & 39.06 & 25.76 & 7.84 & 1.04 & 12.07 \\
Direct+\ours 2x & 83.16 & 68.91	& 17.65 & 55.53 & 54.63 & 27.63 &	3.62 & 29.51 \\
Direct+\ours 4x & 88.95 & \textbf{73.27} & 23.84 & 59.70 & \textbf{62.94} & \textbf{35.44} & \textbf{6.67} & \textbf{35.56} \\
Direct+\ours 6x & \textbf{91.68} & 72.56 & \textbf{24.60} & \textbf{60.59} & 61.92 & 30.91 & 5.39 & 33.61 \\
\bottomrule
\end{tabular}
}
\label{tab:context_limit_abl}
\end{table*} %



\begin{figure}[h]
    \centering
    \includegraphics[width=\linewidth]{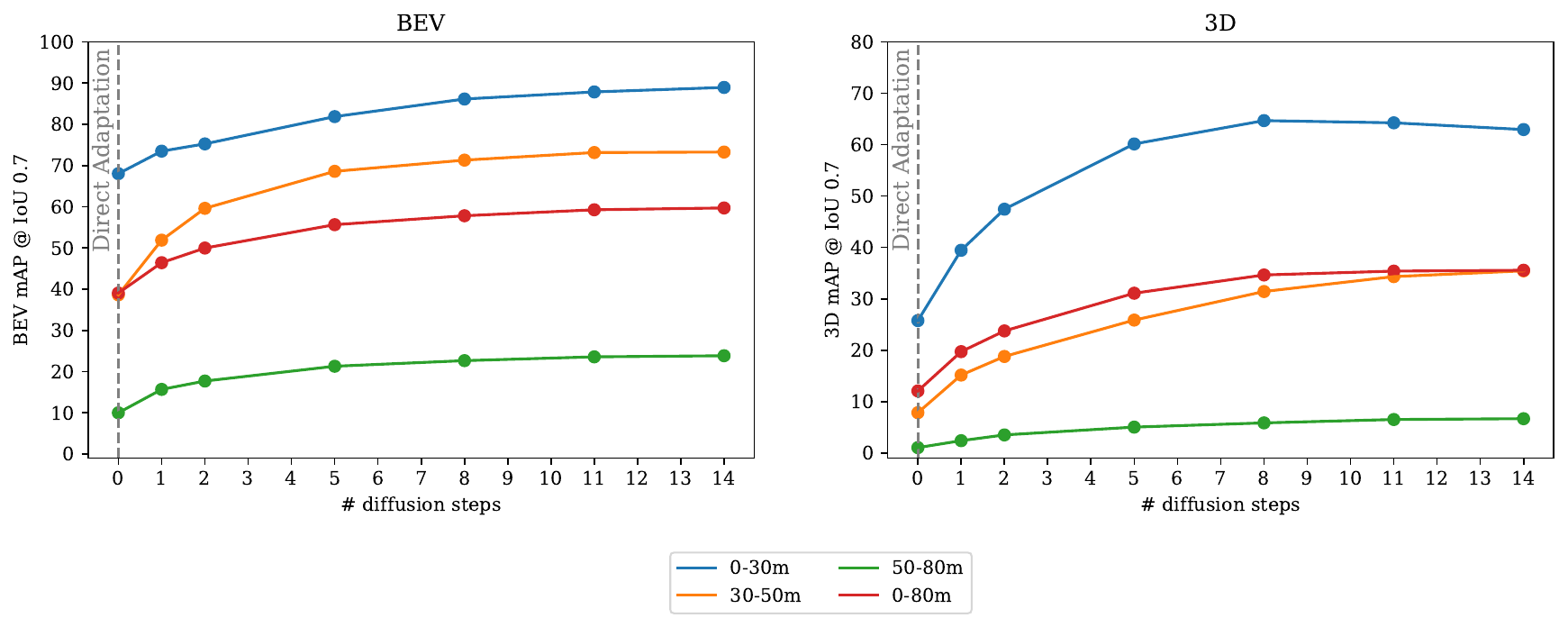}

    \caption{\textbf{mAP vs. Number of Diffusion Steps.} We report the BEV (left) and 3D (right) mAP @ IoU 0.7 for the setting of \kitti $\rightarrow$ \lyft Cars and PointRCNN detector.}
    \label{fig:denoising-steps}
\end{figure}

\paragraph{Denoising Steps.} We perform ablation on the number of denoising steps used, and present the results in \autoref{fig:denoising-steps}. In general, a majority of the performance is already reached using 8 diffusion steps, and it saturates around using 14 steps. 

\paragraph{Shape Weight.} We perform ablation on shape guidance, and consider shape weight 0, 0.01, 0.1 and 0.5. We evaluate under the setting of no adaptation (\textit{\na}) for \kitti $\rightarrow$ \lyft  cars and report the mAP@IoU 0.7 in~\autoref{tab:shape_wt_abl}. \ours improves the detector's output even without shape weight. Using shape weights leads to additional gain, which saturates at around shape weight 0.1. 

\begin{table*}[!t]
\captionsetup{aboveskip=0.5em}
\centering
\caption{\textbf{Ablation on Shape Weight.} \ours improves the detector's output significantly even without using shape weight, and using shape weight attains additional gain.}
  \resizebox{\linewidth}{!}
{

\begin{tabular}{l|cccc|cccc}
\toprule
\multirow{2}{*}{Method} & \multicolumn{4}{c}{BEV$\uparrow$} & \multicolumn{4}{c}{3D$\uparrow$} \\
\cmidrule{2-9}
& \ 0-30m\  & \ 30-50m\  & \ 50-80m\  & \ 0-80m\  & \ 0-30m\  & \ 30-50m\  & \ 50-80m\  & \ 0-80m\  \\
\midrule
Direct & 68.03 & 38.62 & 9.99 & 39.06 & 25.76 & 7.84 & 1.04 & 12.07 \\
No SW & 72.62 & 50.70 & 14.32 & 45.27 & 34.63 & 13.47 & 2.17 & 17.54\\
SW 0.01 & 74.81 & 57.48 & 17.18 & 48.89 & 47.11 & 17.74 & 2.99 & 23.10\\
SW 0.1 & 88.95 & \textbf{73.27} & 23.84 & 59.70 & \textbf{62.94} & \textbf{35.44} & 6.67 & \textbf{35.56}\\
SW 0.5 & \textbf{91.36} & 73.09 & \textbf{24.48} & \textbf{60.23} & 50.70 & 32.73 & \textbf{7.21} & 30.54\\
\bottomrule
\end{tabular}
}
\vspace{-1.2\baselineskip}
\label{tab:shape_wt_abl}
\end{table*} %



\paragraph{IoU Performance Analysis.}
We visualize the comparison of IoU of the bounding boxes with the ground truth bounding boxes both before and after using \ours in \autoref{fig:iou-comp}. The IoU of predictions after \ours refinement (in blue) improves significantly over those before refinement (in yellow). This suggests that a majority of our refinement is in improving the bounding boxes' shape and alignment to fit into the new domain, thus resulting in higher IoU values.

\begin{figure}[ht]
    \centering
    \begin{subfigure}[t]{0.5\textwidth}
        \centering
        \includegraphics[width=\linewidth]{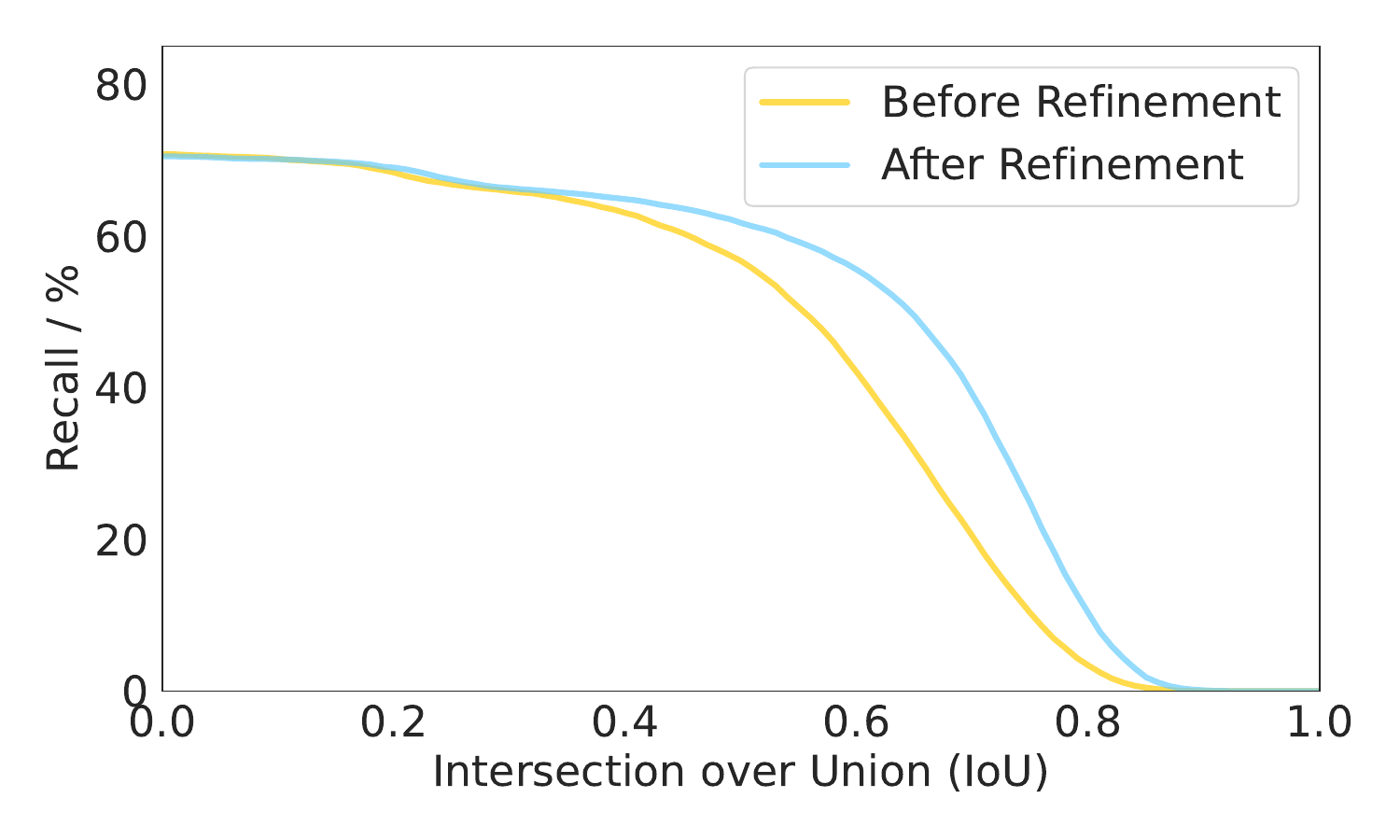}
        \caption{Recall vs. IoU}
        \label{fig:recall_iou}
    \end{subfigure}%
    \begin{subfigure}[t]{0.5\textwidth}
        \centering
        \includegraphics[width=\linewidth]{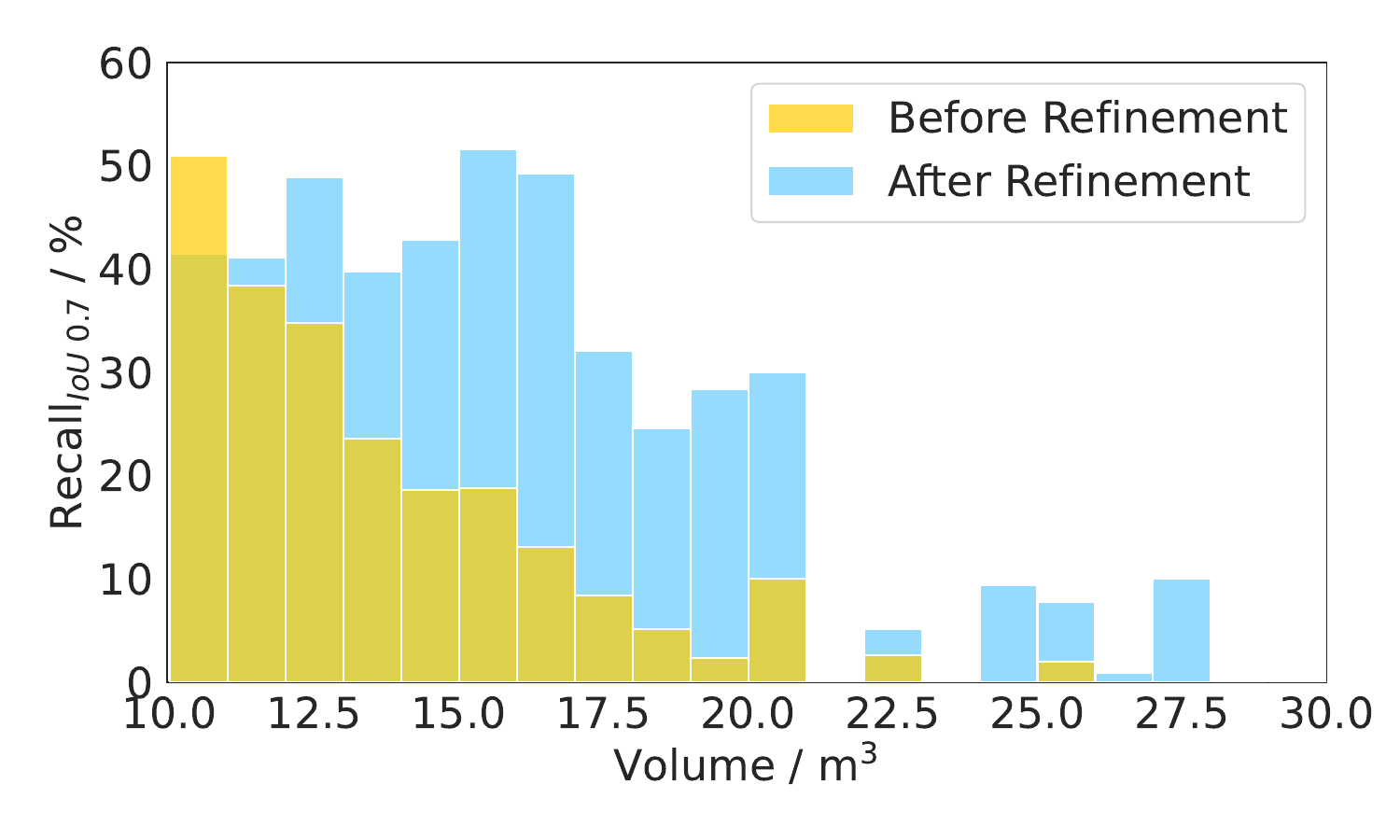}
        \caption{Recall vs. Object volume}
        \label{fig:recall_volume}
    \end{subfigure}%
    \caption{
    \textbf{Recall improvement with \ours.} 
    We report recall on the \emph{car} class (\kitti $\rightarrow$ \lyft, PointRCNN) before and after refinement with \ours.}
    \label{fig:recall}
\end{figure}

\paragraph{Recall Analysis.} We perform analysis on \ours's effect on detection recall in \autoref{fig:recall}. As \ours improves IoU for mislocalized detections, it reduces false negatives that arise from match IoU being lower than the threshold. The improvement is observed for objects across different sizes.

\section{Discussion and Future Work}
In this work, we propose \ours, a diffusion-based approach that refines bounding boxes for better domain adaptation. While \ours effectively improves the existing bounding boxes, one limitation is that \ours currently does not consider false negatives that are due to completely missed detections. This could potentially be addressed through further distilling the detectors by \ours's refined boxes, or by incorporating exploration strategies to capture possibly missed objects. Alternatively, a view that we did not discuss in this work, but is of potential interest to the field, is the use of \ours{} for automatic label refinement. This can be useful for correcting slightly mis-aligned boxes, or labels across sensors that may have slightly unsynchronized sensors. We leave this discussion for further work, and will provide code and model checkpoints for this use case. With our work, we do not foresee any negative societal impacts and hope the field continues to develop such label refinement methods for 3D object detection.


\begin{ack}
This research is supported by grants from the National Science Foundation NSF (IIS-2107161, and IIS-1724282, HDR-2118310). The Cornell Center for Materials Research with funding from the NSF MRSEC program (DMR-1719875), DARPA, arXiv, LinkedIn, Nvidia, and the New York Presbyterian Hospital.
\end{ack}

{\small
\bibliographystyle{plain}
\bibliography{neurips_2024}

\begin{thebibliography}{10}

\bibitem{barnes2018driven}
Dan Barnes, Will Maddern, Geoffrey Pascoe, and Ingmar Posner.
\newblock Driven to distraction: Self-supervised distractor learning for robust monocular visual odometry in urban environments.
\newblock In {\em 2018 IEEE International Conference on Robotics and Automation (ICRA)}, pages 1894--1900. IEEE, 2018.

\bibitem{nuscenes}
Holger Caesar, Varun Bankiti, Alex~H. Lang, Sourabh Vora, Venice~Erin Liong, Qiang Xu, Anush Krishnan, Yu~Pan, Giancarlo Baldan, and Oscar Beijbom.
\newblock nuscenes: A multimodal dataset for autonomous driving.
\newblock In {\em CVPR}, 2020.

\bibitem{chen2023diffusiondet}
Shoufa Chen, Peize Sun, Yibing Song, and Ping Luo.
\newblock Diffusiondet: Diffusion model for object detection.
\newblock In {\em Proceedings of the IEEE/CVF International Conference on Computer Vision}, pages 19830--19843, 2023.

\bibitem{dhariwal2021diffusion}
Prafulla Dhariwal and Alexander Nichol.
\newblock Diffusion models beat gans on image synthesis.
\newblock {\em Advances in neural information processing systems}, 34:8780--8794, 2021.

\bibitem{ithaca365}
Carlos~A Diaz-Ruiz, Youya Xia, Yurong You, Jose Nino, Junan Chen, Josephine Monica, Xiangyu Chen, Katie Luo, Yan Wang, Marc Emond, et~al.
\newblock Ithaca365: Dataset and driving perception under repeated and challenging weather conditions.
\newblock In {\em Proceedings of the IEEE/CVF Conference on Computer Vision and Pattern Recognition}, pages 21383--21392, 2022.

\bibitem{kitti}
Andreas Geiger, Philip Lenz, Christoph Stiller, and Raquel Urtasun.
\newblock Vision meets robotics: The kitti dataset.
\newblock {\em The International Journal of Robotics Research}, 32(11):1231--1237, 2013.

\bibitem{kitti_dataset}
Andreas Geiger, Philip Lenz, and Raquel Urtasun.
\newblock Are we ready for autonomous driving? the kitti vision benchmark suite.
\newblock In {\em Conference on Computer Vision and Pattern Recognition (CVPR)}, 2012.

\bibitem{ho2022imagen}
Jonathan Ho, William Chan, Chitwan Saharia, Jay Whang, Ruiqi Gao, Alexey Gritsenko, Diederik~P. Kingma, Ben Poole, Mohammad Norouzi, David~J. Fleet, and Tim Salimans.
\newblock Imagen video: High definition video generation with diffusion models, 2022.

\bibitem{ho2020denoising}
Jonathan Ho, Ajay Jain, and Pieter Abbeel.
\newblock Denoising diffusion probabilistic models.
\newblock {\em Advances in neural information processing systems}, 33:6840--6851, 2020.

\bibitem{ho2022video}
Jonathan Ho, Tim Salimans, Alexey Gritsenko, William Chan, Mohammad Norouzi, and David~J. Fleet.
\newblock Video diffusion models, 2022.

\bibitem{ince1956ordinary}
Edward~L Ince.
\newblock {\em Ordinary differential equations}.
\newblock Courier Corporation, 1956.

\bibitem{jiang2021lidarnet}
Peng Jiang and Srikanth Saripalli.
\newblock Lidarnet: A boundary-aware domain adaptation model for point cloud semantic segmentation, 2021.

\bibitem{karras2022elucidating}
Tero Karras, Miika Aittala, Timo Aila, and Samuli Laine.
\newblock Elucidating the design space of diffusion-based generative models.
\newblock {\em Advances in Neural Information Processing Systems}, 35:26565--26577, 2022.

\bibitem{edm}
Tero Karras, Miika Aittala, Timo Aila, and Samuli Laine.
\newblock Elucidating the design space of diffusion-based generative models.
\newblock {\em Advances in Neural Information Processing Systems}, 35:26565--26577, 2022.

\bibitem{lyft_dataset}
R.~Kesten, M.~Usman, J.~Houston, T.~Pandya, K.~Nadhamuni, A.~Ferreira, M.~Yuan, B.~Low, A.~Jain, P.~Ondruska, S.~Omari, S.~Shah, A.~Kulkarni, A.~Kazakova, C.~Tao, L.~Platinsky, W.~Jiang, and V.~Shet.
\newblock Lyft level 5 av dataset 2019.
\newblock url{https://level5.lyft.com/dataset/}, 2019.

\bibitem{lyft}
R.~Kesten, M.~Usman, J.~Houston, T.~Pandya, K.~Nadhamuni, A.~Ferreira, M.~Yuan, B.~Low, A.~Jain, P.~Ondruska, S.~Omari, S.~Shah, A.~Kulkarni, A.~Kazakova, C.~Tao, L.~Platinsky, W.~Jiang, and V.~Shet.
\newblock Lyft level 5 av dataset 2019.
\newblock url{https://level5.lyft.com/dataset/}, 2019.

\bibitem{Kim_2023_CVPR}
Hyeonseong Kim, Yoonsu Kang, Changgyoon Oh, and Kuk-Jin Yoon.
\newblock Single domain generalization for lidar semantic segmentation.
\newblock In {\em Proceedings of the IEEE/CVF Conference on Computer Vision and Pattern Recognition (CVPR)}, pages 17587--17598, June 2023.

\bibitem{kim2023diffref3d}
Se-Ho Kim, Inyong Koo, Inyoung Lee, Byeongjun Park, and Changick Kim.
\newblock Diffref3d: A diffusion-based proposal refinement framework for 3d object detection.
\newblock {\em arXiv preprint arXiv:2310.16349}, 2023.

\bibitem{kong2023conda}
Lingdong Kong, Niamul Quader, and Venice~Erin Liong.
\newblock Conda: Unsupervised domain adaptation for lidar segmentation via regularized domain concatenation, 2023.

\bibitem{lang2019pointpillars}
Alex~H Lang, Sourabh Vora, Holger Caesar, Lubing Zhou, Jiong Yang, and Oscar Beijbom.
\newblock Pointpillars: Fast encoders for object detection from point clouds.
\newblock In {\em Proceedings of the IEEE/CVF conference on computer vision and pattern recognition}, pages 12697--12705, 2019.

\bibitem{Langer2020DomainTF}
Ferdinand Langer, Andres Milioto, Alexandre Haag, Jens Behley, and C.~Stachniss.
\newblock Domain transfer for semantic segmentation of lidar data using deep neural networks.
\newblock {\em 2020 IEEE/RSJ International Conference on Intelligent Robots and Systems (IROS)}, pages 8263--8270, 2020.

\bibitem{li2022domain}
Jinlong Li, Runsheng Xu, Jin Ma, Qin Zou, Jiaqi Ma, and Hongkai Yu.
\newblock Domain adaptive object detection for autonomous driving under foggy weather, 2022.

\bibitem{luo2024reward}
Katie Luo, Zhenzhen Liu, Xiangyu Chen, Yurong You, Sagie Benaim, Cheng~Perng Phoo, Mark Campbell, Wen Sun, Bharath Hariharan, and Kilian~Q Weinberger.
\newblock Reward finetuning for faster and more accurate unsupervised object discovery.
\newblock {\em Advances in Neural Information Processing Systems}, 36, 2024.

\bibitem{drift}
Katie~Z Luo, Zhenzhen Liu, Xiangyu Chen, Yurong You, Sagie Benaim, Cheng~Perng Phoo, Mark Campbell, Wen Sun, Bharath Hariharan, and Kilian~Q Weinberger.
\newblock Reward finetuning for faster and more accurate unsupervised object discovery.
\newblock {\em arXiv preprint arXiv:2310.19080}, 2023.

\bibitem{luo2021diffusion}
Shitong Luo and Wei Hu.
\newblock Diffusion probabilistic models for 3d point cloud generation.
\newblock In {\em Proceedings of the IEEE/CVF Conference on Computer Vision and Pattern Recognition}, pages 2837--2845, 2021.

\bibitem{mao2021voxel}
Jiageng Mao, Yujing Xue, Minzhe Niu, Haoyue Bai, Jiashi Feng, Xiaodan Liang, Hang Xu, and Chunjing Xu.
\newblock Voxel transformer for 3d object detection.
\newblock In {\em Proceedings of the IEEE/CVF International Conference on Computer Vision}, pages 3164--3173, 2021.

\bibitem{Melas-Kyriazi_2023_CVPR}
Luke Melas-Kyriazi, Christian Rupprecht, and Andrea Vedaldi.
\newblock Pc2: Projection-conditioned point cloud diffusion for single-image 3d reconstruction.
\newblock In {\em Proceedings of the IEEE/CVF Conference on Computer Vision and Pattern Recognition (CVPR)}, pages 12923--12932, June 2023.

\bibitem{morerio2017minimalentropy}
Pietro Morerio, Jacopo Cavazza, and Vittorio Murino.
\newblock Minimal-entropy correlation alignment for unsupervised deep domain adaptation, 2017.

\bibitem{pan20213d}
Xuran Pan, Zhuofan Xia, Shiji Song, Li~Erran Li, and Gao Huang.
\newblock 3d object detection with pointformer.
\newblock In {\em Proceedings of the IEEE/CVF Conference on Computer Vision and Pattern Recognition}, pages 7463--7472, 2021.

\bibitem{poole2022dreamfusion}
Ben Poole, Ajay Jain, Jonathan~T Barron, and Ben Mildenhall.
\newblock Dreamfusion: Text-to-3d using 2d diffusion.
\newblock {\em arXiv preprint arXiv:2209.14988}, 2022.

\bibitem{qi2018frustum}
Charles~R Qi, Wei Liu, Chenxia Wu, Hao Su, and Leonidas~J Guibas.
\newblock Frustum pointnets for 3d object detection from rgb-d data.
\newblock In {\em Proceedings of the IEEE conference on computer vision and pattern recognition}, pages 918--927, 2018.

\bibitem{qi2017pointnet}
Charles~R Qi, Hao Su, Kaichun Mo, and Leonidas~J Guibas.
\newblock Pointnet: Deep learning on point sets for 3d classification and segmentation.
\newblock In {\em Proceedings of the IEEE conference on computer vision and pattern recognition}, pages 652--660, 2017.

\bibitem{qi2017pointnet++}
Charles~Ruizhongtai Qi, Li~Yi, Hao Su, and Leonidas~J Guibas.
\newblock Pointnet++: Deep hierarchical feature learning on point sets in a metric space.
\newblock {\em Advances in neural information processing systems}, 30, 2017.

\bibitem{latentdiffusion}
Robin Rombach, Andreas Blattmann, Dominik Lorenz, Patrick Esser, and Bj{\"o}rn Ommer.
\newblock High-resolution image synthesis with latent diffusion models.
\newblock In {\em Proceedings of the IEEE/CVF conference on computer vision and pattern recognition}, pages 10684--10695, 2022.

\bibitem{saito2018adversarial}
Kuniaki Saito, Yoshitaka Ushiku, Tatsuya Harada, and Kate Saenko.
\newblock Adversarial dropout regularization, 2018.

\bibitem{saleh2019domain}
Khaled Saleh, Ahmed Abobakr, Mohammed Attia, Julie Iskander, Darius Nahavandi, and Mohammed Hossny.
\newblock Domain adaptation for vehicle detection from bird's eye view lidar point cloud data, 2019.

\bibitem{saltori2022cosmix}
Cristiano Saltori, Fabio Galasso, Giuseppe Fiameni, Nicu Sebe, Elisa Ricci, and Fabio Poiesi.
\newblock Cosmix: Compositional semantic mix for domain adaptation in 3d lidar segmentation, 2022.

\bibitem{shi2020pv}
Shaoshuai Shi, Chaoxu Guo, Li~Jiang, Zhe Wang, Jianping Shi, Xiaogang Wang, and Hongsheng Li.
\newblock Pv-rcnn: Point-voxel feature set abstraction for 3d object detection.
\newblock In {\em Proceedings of the IEEE/CVF conference on computer vision and pattern recognition}, pages 10529--10538, 2020.

\bibitem{shi2019pointrcnn}
Shaoshuai Shi, Xiaogang Wang, and Hongsheng Li.
\newblock Pointrcnn: 3d object proposal generation and detection from point cloud.
\newblock In {\em Proceedings of the IEEE/CVF conference on computer vision and pattern recognition}, pages 770--779, 2019.

\bibitem{shi2020point}
Weijing Shi and Raj Rajkumar.
\newblock Point-gnn: Graph neural network for 3d object detection in a point cloud.
\newblock In {\em Proceedings of the IEEE/CVF conference on computer vision and pattern recognition}, pages 1711--1719, 2020.

\bibitem{sohl2015deep}
Jascha Sohl-Dickstein, Eric Weiss, Niru Maheswaranathan, and Surya Ganguli.
\newblock Deep unsupervised learning using nonequilibrium thermodynamics.
\newblock In {\em International conference on machine learning}, pages 2256--2265. PMLR, 2015.

\bibitem{song2020denoising}
Jiaming Song, Chenlin Meng, and Stefano Ermon.
\newblock Denoising diffusion implicit models.
\newblock {\em arXiv preprint arXiv:2010.02502}, 2020.

\bibitem{song2020score}
Yang Song, Jascha Sohl-Dickstein, Diederik~P Kingma, Abhishek Kumar, Stefano Ermon, and Ben Poole.
\newblock Score-based generative modeling through stochastic differential equations.
\newblock {\em arXiv preprint arXiv:2011.13456}, 2020.

\bibitem{openpcdet2020}
OpenPCDet~Development Team.
\newblock Openpcdet: An open-source toolbox for 3d object detection clouds.
\newblock \url{https://github.com/open-mmlab/OpenPCDet}, 2020.

\bibitem{van1976stochastic}
Nicolaas~G Van~Kampen.
\newblock Stochastic differential equations.
\newblock {\em Physics reports}, 24(3):171--228, 1976.

\bibitem{vu2019advent}
Tuan-Hung Vu, Himalaya Jain, Maxime Bucher, Matthieu Cord, and Patrick Pérez.
\newblock Advent: Adversarial entropy minimization for domain adaptation in semantic segmentation, 2019.

\bibitem{wang2023dsvt}
Haiyang Wang, Chen Shi, Shaoshuai Shi, Meng Lei, Sen Wang, Di~He, Bernt Schiele, and Liwei Wang.
\newblock Dsvt: Dynamic sparse voxel transformer with rotated sets.
\newblock In {\em Proceedings of the IEEE/CVF Conference on Computer Vision and Pattern Recognition}, pages 13520--13529, 2023.

\bibitem{wang2020train}
Yan Wang, Xiangyu Chen, Yurong You, Li~Erran Li, Bharath Hariharan, Mark Campbell, Kilian~Q Weinberger, and Wei-Lun Chao.
\newblock Train in germany, test in the usa: Making 3d object detectors generalize.
\newblock In {\em Proceedings of the IEEE/CVF Conference on Computer Vision and Pattern Recognition}, pages 11713--11723, 2020.

\bibitem{wang2020pillar}
Yue Wang, Alireza Fathi, Abhijit Kundu, David~A Ross, Caroline Pantofaru, Tom Funkhouser, and Justin Solomon.
\newblock Pillar-based object detection for autonomous driving.
\newblock In {\em Computer Vision--ECCV 2020: 16th European Conference, Glasgow, UK, August 23--28, 2020, Proceedings, Part XXII 16}, pages 18--34. Springer, 2020.

\bibitem{xu2021squeezesegv3}
Chenfeng Xu, Bichen Wu, Zining Wang, Wei Zhan, Peter Vajda, Kurt Keutzer, and Masayoshi Tomizuka.
\newblock Squeezesegv3: Spatially-adaptive convolution for efficient point-cloud segmentation, 2021.

\bibitem{yan2018second}
Yan Yan, Yuxing Mao, and Bo~Li.
\newblock Second: Sparsely embedded convolutional detection.
\newblock {\em Sensors}, 18(10):3337, 2018.

\bibitem{yang2019pointflow}
Guandao Yang, Xun Huang, Zekun Hao, Ming-Yu Liu, Serge Belongie, and Bharath Hariharan.
\newblock Pointflow: 3d point cloud generation with continuous normalizing flows.
\newblock In {\em Proceedings of the IEEE/CVF international conference on computer vision}, pages 4541--4550, 2019.

\bibitem{st3d}
Jihan Yang, Shaoshuai Shi, Zhe Wang, Hongsheng Li, and Xiaojuan Qi.
\newblock St3d: Self-training for unsupervised domain adaptation on 3d object detection.
\newblock In {\em Proceedings of the IEEE/CVF conference on computer vision and pattern recognition}, pages 10368--10378, 2021.

\bibitem{yang2022st3d++}
Jihan Yang, Shaoshuai Shi, Zhe Wang, Hongsheng Li, and Xiaojuan Qi.
\newblock St3d++: Denoised self-training for unsupervised domain adaptation on 3d object detection.
\newblock {\em IEEE transactions on pattern analysis and machine intelligence}, 45(5):6354--6371, 2022.

\bibitem{yang20203dssd}
Zetong Yang, Yanan Sun, Shu Liu, and Jiaya Jia.
\newblock 3dssd: Point-based 3d single stage object detector.
\newblock In {\em Proceedings of the IEEE/CVF conference on computer vision and pattern recognition}, pages 11040--11048, 2020.

\bibitem{yin2021center}
Tianwei Yin, Xingyi Zhou, and Philipp Krahenbuhl.
\newblock Center-based 3d object detection and tracking.
\newblock In {\em Proceedings of the IEEE/CVF conference on computer vision and pattern recognition}, pages 11784--11793, 2021.

\bibitem{you2022exploiting}
Yurong You, Carlos~Andres Diaz-Ruiz, Yan Wang, Wei-Lun Chao, Bharath Hariharan, Mark Campbell, and Kilian~Q Weinbergert.
\newblock Exploiting playbacks in unsupervised domain adaptation for 3d object detection in self-driving cars.
\newblock In {\em 2022 International Conference on Robotics and Automation (ICRA)}, pages 5070--5077. IEEE, 2022.

\bibitem{modest}
Yurong You, Katie Luo, Cheng~Perng Phoo, Wei-Lun Chao, Wen Sun, Bharath Hariharan, Mark Campbell, and Kilian~Q Weinberger.
\newblock Learning to detect mobile objects from lidar scans without labels.
\newblock In {\em Proceedings of the IEEE/CVF Conference on Computer Vision and Pattern Recognition}, pages 1130--1140, 2022.

\bibitem{roteda}
Yurong You, Cheng~Perng Phoo, Katie Luo, Travis Zhang, Wei-Lun Chao, Bharath Hariharan, Mark Campbell, and Kilian~Q Weinberger.
\newblock Unsupervised adaptation from repeated traversals for autonomous driving.
\newblock {\em Advances in Neural Information Processing Systems}, 35:27716--27729, 2022.

\bibitem{you2022unsupervised}
Yurong You, Cheng~Perng Phoo, Katie Luo, Travis Zhang, Wei-Lun Chao, Bharath Hariharan, Mark Campbell, and Kilian~Q Weinberger.
\newblock Unsupervised adaptation from repeated traversals for autonomous driving.
\newblock {\em Advances in Neural Information Processing Systems}, 35:27716--27729, 2022.

\bibitem{zeng2022lion}
Xiaohui Zeng, Arash Vahdat, Francis Williams, Zan Gojcic, Or~Litany, Sanja Fidler, and Karsten Kreis.
\newblock Lion: Latent point diffusion models for 3d shape generation.
\newblock {\em arXiv preprint arXiv:2210.06978}, 2022.

\bibitem{Zhou_2021_ICCV}
Linqi Zhou, Yilun Du, and Jiajun Wu.
\newblock 3d shape generation and completion through point-voxel diffusion.
\newblock In {\em Proceedings of the IEEE/CVF International Conference on Computer Vision (ICCV)}, pages 5826--5835, October 2021.

\bibitem{zhou20213d}
Linqi Zhou, Yilun Du, and Jiajun Wu.
\newblock 3d shape generation and completion through point-voxel diffusion.
\newblock In {\em Proceedings of the IEEE/CVF international conference on computer vision}, pages 5826--5835, 2021.

\bibitem{zhou2023diffusion}
Xin Zhou, Jinghua Hou, Tingting Yao, Dingkang Liang, Zhe Liu, Zhikang Zou, Xiaoqing Ye, Jianwei Cheng, and Xiang Bai.
\newblock Diffusion-based 3d object detection with random boxes.
\newblock {\em arXiv preprint arXiv:2309.02049}, 2023.

\bibitem{zhou2018voxelnet}
Yin Zhou and Oncel Tuzel.
\newblock Voxelnet: End-to-end learning for point cloud based 3d object detection.
\newblock In {\em Proceedings of the IEEE conference on computer vision and pattern recognition}, pages 4490--4499, 2018.

\end{thebibliography}
}


\appendix


\vbox{

\centering
    {\LARGE\bf 
    Supplementary Material: \\
    \ourslong
    \par}
}

\appendix
\renewcommand{\thesection}{S\arabic{section}}  
\renewcommand{\thetable}{S\arabic{table}}  
\renewcommand{\thefigure}{S\arabic{figure}}


\section{Algorithmic Description of \ours}\label{sec:alg-desc}
Below we provide an algorithmic description of the training and inference workflow of \ours.

\begin{minipage}[t]{0.42\linewidth}
\begin{algorithm}[H]
\caption{\ours{} Training}
\begin{algorithmic}[1]
\Require Train set $\mathcal{D}_{\text{train}}$, noise distribution $p_{\text{train}}$
\Ensure Diffusion model $F_\theta$ to approximate $\nabla_{\boldsymbol{x}} \log p(\boldsymbol{x}, \sigma)$ of bounding box over points $\boldsymbol{x}$, noise $\sigma$
\Repeat
    \State Sample point cloud $\boldsymbol{P}$, true bounding box $\boldsymbol{b^*}$ where $\boldsymbol{P}, \boldsymbol{b^*} \sim \mathcal{D}_{\text{train}}$
    \State Sample noise level $\sigma \sim p_{\text{train}}$
    \State Compute $\boldsymbol{P}_{\boldsymbol{b^*}}^{\text{NBV}}$ from Eq. 2
    \State Update $F_\theta$ with diffusion objective Eq. 4
\Until convergence
\end{algorithmic}
\end{algorithm}
\end{minipage}%
\hfill
\begin{minipage}[t]{0.55\linewidth}
\begin{algorithm}[H]
\caption{\ours{} Inference}
\begin{algorithmic}[1]
\Require Set of initial bounding boxes $\{\boldsymbol{b}\}_{\text{scene}}$, point cloud $\boldsymbol{P}$, $F_\theta$ approximating $\nabla_{\boldsymbol{x}} \log p(\boldsymbol{x}, \sigma)$
\Ensure Set of refined bounding boxes $\{\hat{\boldsymbol{b}}\}_{\text{scene}}$
\For{each bounding box $\boldsymbol{b}$ in $\{\boldsymbol{b}\}_{\text{scene}}$}
    \For{denoising steps $t = 1, \cdots, T$}
        \State Obtain $\sigma(t)$ from denoising schedule
        \State Compute $\boldsymbol{P}_{\boldsymbol{b}}^{\text{NBV}}$ from Eq. 2
        \State Compute $\text{d} \boldsymbol{b}$ with $\sigma(t)$ and $F_\theta$ from Eq. 8
        \State $\boldsymbol{b} \gets \boldsymbol{b} + \text{d} \boldsymbol{b}$
    \EndFor
    \State $\hat{\boldsymbol{b}} \gets \boldsymbol{b}$
\EndFor
\State $\{\hat{\boldsymbol{b}}\}_{\text{scene}} \gets \text{NMS}(\{\hat{\boldsymbol{b}}\}_{\text{scene}})$
\end{algorithmic}
\end{algorithm}
\end{minipage}

\section{Implementation Details}

\paragraph{Experimental Setup.} For detectors, we use the implementation and configurations from OpenPCDet~\cite{openpcdet2020}. For diffusion models, we use \cite{edm}'s implementation and follow their noise schedule $\sigma_{max}=80$. During inference, we denoise each box for 14 steps, starting from a noise level between $[10,80]$ linear in the detector confidence; higher confidence boxes begin from smaller noise. For each bounding box, we consider the points that fall within the range of four times its size. Additionally, inspired by OT and SN, we slightly regularize the bounding box shape with the average size of the target domain, with shape weight 0.1 for cars and pedestrians, and 0.01 for cyclists. We use NVIDIA A6000 for all of our experiments.

\begin{figure*}[h]
    \centering
    \includegraphics[width=.9\textwidth]{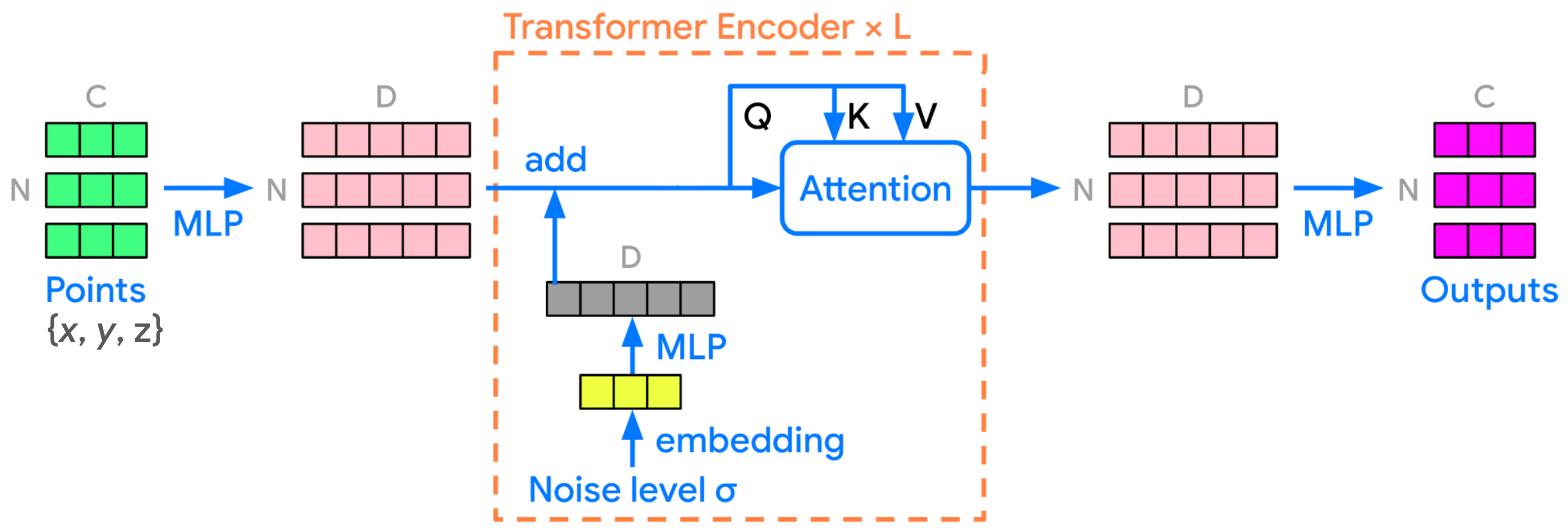}
    \caption{
    Architecture overview of \ours's denoiser model. 
    The model is composed of 2 MLP layers and $L$ transformer encoder layers, which maps 3D points to a higher dimensional space for effective self-attention.
    }
    \label{fig:architecture}
\end{figure*}

\paragraph{Training Details.}
In the diffusion process, we follow \cite{karras2022elucidating} and use noise level distribution $\ln \sigma \sim \mathcal{N}\left(-1.2, {1.2}^2\right)$, ODE schedule $\sigma(t) = t$, and $2^{\text{nd}}$ order Heun solver.
The denoiser transformer model contains 12 self-attention layers with hidden size 1024; each layer has 2048 intermediate dimensions and 8 heads.
The diffusion model is trained with batch size 128 and learning rate 0.0001 for 100k steps. A visualization of our Diffusion model architecture can be found in \autoref{fig:architecture}.


\section{Additional Experiment Results}

\subsection{\nuscenes True Positive Metrics}

We report the \nuscenes TP metrics for \kitti $\rightarrow$ \lyft cars in \autoref{tab:nuscenes_tp}, and those for \kitti $\rightarrow$ \ith in \autoref{tab:nuscenes_tp_ith}. All results are based on PointRCNN detectors. \ours effectively reduces all three types of errors, with especially significant improvement for translation and scale errors.   

\begin{table*}[!htbp]
\captionsetup{aboveskip=0.5em}
\centering
\caption{\textbf{\nuscenes TP metrics for \kitti $\rightarrow$ \lyft (cars).} Lower is better. \ours reduces all three types of errors, with especially significant improvement for translation and scale errors. }
  \resizebox{\linewidth}{!}
{

\begin{tabular}{l|cccc|cccc|cccc}
\toprule
\multirow{2}{*}{Method} & \multicolumn{4}{c}{Translation Error$\downarrow$} & \multicolumn{4}{c}{Scale Error$\downarrow$} & \multicolumn{4}{c}{Orientation Error$\downarrow$}\\
\cmidrule{2-13}
& \ 0-30m\  & \ 30-50m\  & \ 50-80m\  & \ 0-80m\  & \ 0-30m\  & \ 30-50m\  & \ 50-80m\  & \ 0-80m\ & \ 0-30m\ & \ 30-50m\  & \ 50-80m\  & \ 0-80m\  \\
\midrule
Direct & 0.346 & 0.473 & 0.506 & 0.430 & 0.267 & 0.299 & 0.309 & 0.289 & 0.347 & 0.510 & 0.624 & 0.472 \\
Direct+\ours & \textbf{0.221} & \textbf{0.289} & \textbf{0.424} & \textbf{0.293} & \textbf{0.162} & \textbf{0.171} & \textbf{0.198} & \textbf{0.173} & \textbf{0.329} & \textbf{0.473} & \textbf{0.607} & \textbf{0.447}\\\midrule
OT & 0.273 & 0.376 & 0.466 & 0.356 & 0.217 & 0.207 & 0.219 & 0.214 & 0.338 & 0.496 & 0.623 & 0.463 \\
OT+\ours & \textbf{0.201} & \textbf{0.266} & \textbf{0.399} & \textbf{0.271} & \textbf{0.174} & \textbf{0.173} & \textbf{0.202} & \textbf{0.180} & \textbf{0.328} & \textbf{0.474} & \textbf{0.598} & \textbf{0.444} \\\midrule
SN & 0.332 & 0.436 & 0.440 & 0.398 & \textbf{0.160} & 0.178 & \textbf{0.204} & \textbf{0.178} & 0.320 & 0.540 & 0.678 & 0.493\\
SN+\ours & \textbf{0.201} & \textbf{0.251} & \textbf{0.410} & \textbf{0.274} & 0.164 & \textbf{0.171} & 0.207 & \textbf{0.178} & \textbf{0.309} & \textbf{0.505} & \textbf{0.631} & \textbf{0.463} \\\midrule
Rote-DA & 0.286 & 0.352 & \textbf{0.408} & 0.338 & 0.203 & 0.213 & \textbf{0.190} & 0.204 & 0.253 & 0.448 & 0.621 & 0.408 \\
Rote-DA+\ours & \textbf{0.206} & \textbf{0.260} & 0.423 & \textbf{0.275} & \textbf{0.161} & \textbf{0.170} & 0.201 & \textbf{0.173} & \textbf{0.244} & \textbf{0.437} & \textbf{0.609} & \textbf{0.398} \\\midrule
ST3D & 0.386 & 0.489 & 0.451 & 0.440 & 0.240 & 0.225 & 0.239 & 0.234 & 0.352 & 0.562 & 0.693 & 0.517\\
ST3D+\ours & \textbf{0.216} & \textbf{0.262} & \textbf{0.350} & \textbf{0.267} & \textbf{0.162} & \textbf{0.171} & \textbf{0.199} & \textbf{0.175} & \textbf{0.326} & \textbf{0.508} & \textbf{0.681} & \textbf{0.485} \\
\bottomrule
\end{tabular}
}
\label{tab:nuscenes_tp}
\end{table*}

\begin{table*}[!htbp]
\captionsetup{aboveskip=0.5em}
\centering
\caption{\textbf{\nuscenes TP metrics for \kitti $\rightarrow$ \ith (cars).} Lower is better. }
  \resizebox{\linewidth}{!}
{

\begin{tabular}{l|cccc|cccc|cccc}
\toprule
\multirow{2}{*}{Method} & \multicolumn{4}{c}{Translation Error$\downarrow$} & \multicolumn{4}{c}{Scale Error$\downarrow$} & \multicolumn{4}{c}{Orientation Error$\downarrow$}\\
\cmidrule{2-13}
& \ 0-30m\  & \ 30-50m\  & \ 50-80m\  & \ 0-80m\  & \ 0-30m\  & \ 30-50m\  & \ 50-80m\  & \ 0-80m\ & \ 0-30m\ & \ 30-50m\  & \ 50-80m\  & \ 0-80m\  \\
\midrule
Direct & 0.403 & 0.523 & \textbf{0.722} & 0.510 & 0.200 & 0.206 & 0.207 & 0.204 & \textbf{0.492} & \textbf{0.827} & \textbf{1.001} &\textbf{0.705}\\
Direct+\ours & \textbf{0.312} & \textbf{0.456} & 0.775 & \textbf{0.458} & \textbf{0.125} & \textbf{0.131} & \textbf{0.172} & \textbf{0.137} & 0.493 & 0.831 & 1.002 & 0.708\\\midrule
OT & 0.381 & 0.506 & \textbf{0.720} & 0.494 & 0.159 & 0.150 & \textbf{0.139} & 0.152 & \textbf{0.492} & \textbf{0.827} & 1.004 & 0.707 \\
OT+\ours & \textbf{0.320} & \textbf{0.458} & 0.753 & \textbf{0.456} & \textbf{0.124} & \textbf{0.133} & 0.173 & \textbf{0.138} & 0.494 & 0.829 & \textbf{0.997} & \textbf{0.705} \\\midrule
SN & 0.484 & 0.544 & \textbf{0.816} & 0.587 & 0.153 & 0.141 & \textbf{0.150} & 0.149 & 0.590	& 0.869 & 1.091 & 0.800\\
SN+\ours & \textbf{0.351} & \textbf{0.492} & 0.862 & \textbf{0.520} & \textbf{0.128} & \textbf{0.137} & 0.179 & \textbf{0.144} & \textbf{0.574} & \textbf{0.862} & \textbf{1.072} & \textbf{0.783}\\\midrule
Rote-DA & 0.492 & 0.484 & \textbf{0.461} & 0.482 & 0.140 & 0.144 & 0.167 & 0.147 & 0.476 & 0.731 & 1.023 & 0.683 \\
Rote-DA+\ours & \textbf{0.315} & \textbf{0.384} & 0.517 & \textbf{0.383} & \textbf{0.121} & \textbf{0.116} & \textbf{0.135} & \textbf{0.123} & \textbf{0.451} & \textbf{0.715} & \textbf{0.984} & \textbf{0.656} \\
\bottomrule
\end{tabular}
}
\label{tab:nuscenes_tp_ith}
\end{table*}

\subsection{Additional Experimental Results}
We report the performance on \kitti $\rightarrow$ \nuscenes cars in \autoref{tab:adaptation_car_nusc}, and \kitti $\rightarrow$ \ith pedestrians in \autoref{tab:adaptation_pedestrian_ith}. Both sets of experiments are based on PointRCNN detectors. For \nuscenes, we evaluate for up to 50m based on the point cloud range. \ours attains significant improvement upon the detections in both settings.
\begin{table*}[!htbp]
\captionsetup{aboveskip=0.5em}
\centering
\caption{\textbf{mAP@IoU 0.7 for \kitti $\rightarrow$ \nuscenes (cars).} \ours consistently attains significant improvement. }
  \resizebox{0.8\linewidth}{!}
{

\begin{tabular}{l|ccc|ccc}
\toprule
 \multirow{2}{*}{Method} & \multicolumn{3}{c}{BEV $\uparrow$} & \multicolumn{3}{c}{3D $\uparrow$} \\
\cmidrule{2-7}
& \ 0-30m\  & \ 30-50m\   & \ 0-50m\  & \ 0-30m\  & \ 30-50m\   & \ 0-50m\  \\
\midrule
 Direct & 44.78 & 0.70 & 15.86 & 14.82 & 0.00 & 4.66\\
 Direct+\ours & \textbf{58.07} & \textbf{1.06} & \textbf{20.70} & \textbf{22.77} & 0.00 & \textbf{7.40} \\
\bottomrule
\end{tabular}
}
\label{tab:adaptation_car_nusc}
\end{table*} %

\begin{table*}[!htbp]
\captionsetup{aboveskip=0.5em}
\centering
\caption{\textbf{mAP@IoU 0.5 for \kitti $\rightarrow$ \ith (pedestrians).} \ours consistently attains significant improvement. }
  \resizebox{\linewidth}{!}
{

\begin{tabular}{l|cccc|cccc}
\toprule
 \multirow{2}{*}{Method} & \multicolumn{4}{c}{BEV $\uparrow$} & \multicolumn{4}{c}{3D $\uparrow$} \\
\cmidrule{2-9}
& \ 0-30m\  & \ 30-50m\  & \ 50-80m\  & \ 0-80m\  & \ 0-30m\  & \ 30-50m\  & \ 50-80m\  & \ 0-80m\  \\
\midrule
 Direct & 41.70 & 16.87 & \textbf{1.63} & 23.07 & 31.38 & 10.06 & \textbf{0.60} & 16.09 \\
 Direct+\ours & \textbf{50.53} & \textbf{20.26} & 0.49 & \textbf{26.62} & \textbf{43.36} & \textbf{14.74} & 0.10 & \textbf{21.74}\\\midrule
 OT &  42.96 & 18.66 & \textbf{1.75} & 23.69 & 33.00 & 11.64 & \textbf{0.63} & 17.15\\
OT+\ours & \textbf{50.52} & \textbf{21.27} & 0.38 & \textbf{26.95} & \textbf{44.67} & \textbf{15.45} & 0.05 & \textbf{22.48} \\\midrule
SN & 48.95 & 16.97 & \textbf{2.46} & 26.36 & 38.47 & 9.17 & \textbf{0.97} & 17.89\\
SN+\ours & \textbf{57.32} & \textbf{17.27} & 0.51 & \textbf{28.20} & \textbf{47.63} & \textbf{10.44} & 0.22 & \textbf{21.62} \\\midrule
Rote-DA & 43.36 & 1.68 & 0.00 & 14.18 & 30.21 & 0.13 & 0.00 & 8.40 \\
Rote-DA+\ours & \textbf{56.91} & \textbf{21.31} & \textbf{1.18} & \textbf{29.75} & \textbf{50.94} & \textbf{14.63} & \textbf{0.31} & \textbf{24.18} \\
\bottomrule
\end{tabular}
}
\label{tab:adaptation_pedestrian_ith}
\end{table*} %

In addition, we report the full \kitti $\rightarrow$ \lyft performance on all three traffic participant classes with the CenterPoint~\citep{yin2021center} detector in \autoref{tab:centerpoint} and DSVT~\citep{wang2023dsvt} in \autoref{tab:dsvt}. \ours achieves consistent improvement across object detectors, classes, and domain adaptation methods.
\begin{table*}[!th]
\captionsetup{aboveskip=0.5em}
\centering
\caption{\textbf{mAP for \kitti $\rightarrow$ \lyft with CenterPoint~\citep{yin2021center}.} \ours attains consistent improvement across object classes and domain adaptation methods. }
  \resizebox{\linewidth}{!}
{

\begin{tabular}{l|l|cccc|cccc}
\toprule
\multirow{2}{*}{ Class } & \multirow{2}{*}{ Method } & \multicolumn{4}{c}{BEV $\uparrow$} & \multicolumn{4}{c}{3D $\uparrow$} \\
\cmidrule{3-10}
& & 0-30m & 30-50m & 50-80m & 0-80m & 0-30m & 30-50m & 50-80m & 0-80m \\
\midrule
\multirow{4}{*}{ Car@IoU 0.7 } & Direct & 74.91 & 36.64 & 2.47 & 36.23 & 28.63 & 4.05 & 0.15 & 10.29 \\
& Direct+DiffuBox & \textbf{90.25} & \textbf{58.65} & \textbf{6.95} & \textbf{51.38} & \textbf{71.02} & \textbf{34.91} & \textbf{1.48} & \textbf{34.40} \\
\cmidrule{2-10}
& OT & 82.81 & 51.02 & 6.26 & 46.10 & 25.34 & 12.35 & 0.52 & 13.34 \\
& OT+DiffuBox & \textbf{91.79} & \textbf{59.24} & \textbf{7.45} & \textbf{51.88} & \textbf{63.01} & \textbf{32.53} & \textbf{1.24} & \textbf{30.95} \\
\midrule
\multirow{4}{*}{ Pedestrian@IoU 0.5 } & Direct & 0.85 & 1.22 & 0.02 & 0.49 & 0.41 & 0.47 & 0.01 & 0.19 \\
& Direct+DiffuBox & \textbf{5.58} & \textbf{2.06} & \textbf{0.03} & \textbf{1.78} & \textbf{3.99} & \textbf{1.75} & \textbf{0.02} & \textbf{1.20} \\
\cmidrule{2-10}
& OT & 1.88 & 2.02 & \textbf{0.05} & 0.87 & 0.61 & 1.22 & 0.01 & 0.37 \\
& OT+DiffuBox & \textbf{5.78} & \textbf{2.32} & 0.03 & \textbf{1.78} & \textbf{4.04} & \textbf{1.90} & \textbf{0.02} & \textbf{1.16} \\
\midrule
\multirow{4}{*}{ Cyclist@IoU 0.5 } & Direct & 23.34 & 0.84 & 0.01 & 8.19 & 16.53 & 0.34 & 0.00 & 5.42 \\
& Direct+DiffuBox & \textbf{42.22} & \textbf{1.49} & \textbf{0.02} & \textbf{15.22} & \textbf{31.74} & \textbf{0.58} & \textbf{0.01} & \textbf{11.28} \\
\cmidrule{2-10}
& OT & 27.40 & 0.95 & 0.01 & 9.73 & 7.32 & 0.05 & \textbf{0.01} & 1.98 \\
& OT+DiffuBox & \textbf{42.22} & \textbf{1.54} & \textbf{0.02} & \textbf{15.23} & \textbf{31.35} & \textbf{0.67} & \textbf{0.01} & \textbf{11.20} \\
\bottomrule
\end{tabular}
}
\label{tab:centerpoint}
\end{table*} %

\begin{table*}[!th]
\captionsetup{aboveskip=0.5em}
\centering
\caption{\textbf{mAP for \kitti $\rightarrow$ \lyft with DSVT~\citep{wang2023dsvt}.} \ours attains consistent improvement across object classes and domain adaptation methods. }
  \resizebox{\linewidth}{!}
{

\begin{tabular}{l|l|cccc|cccc}
\toprule
\multirow{2}{*}{ Class } & \multirow{2}{*}{ Method } & \multicolumn{4}{c}{BEV $\uparrow$} & \multicolumn{4}{c}{3D $\uparrow$} \\
\cmidrule{3-10}
& & 0-30m & 30-50m & 50-80m & 0-80m & 0-30m & 30-50m & 50-80m & 0-80m \\
\midrule
\multirow{4}{*}{ Car@IoU 0.7 } & Direct & 68.93 & 47.49 & 11.32 & 41.77 & 33.72 & 11.74 & 1.42 & 15.67 \\
& Direct+\ours & \textbf{89.01} & \textbf{63.41} & \textbf{17.50} & \textbf{55.27} & \textbf{65.22} & \textbf{36.31} & \textbf{5.02} & \textbf{35.61} \\
\cmidrule{2-10}
& OT & 71.85 & 42.93 & 13.18 & 43.05 & 15.66 & 4.47 & 0.31 & 8.06 \\
& OT+\ours & \textbf{90.21} & \textbf{63.50} & \textbf{17.91} & \textbf{55.61} & \textbf{56.84} & \textbf{31.19} & \textbf{4.63} & \textbf{31.12} \\
\midrule
\multirow{4}{*}{ Pedestrian@IoU 0.5 } & Direct & 16.42 & 6.48 & 0.47 & 8.09 & 11.49 & 4.41 & 0.10 & 5.32 \\
& Direct+\ours & \textbf{27.89} & \textbf{8.13} & \textbf{1.05} & \textbf{12.42} & \textbf{23.55} & \textbf{6.49} & \textbf{0.53} & \textbf{10.28} \\
\cmidrule{2-10}
& OT & 20.89 & 7.47 & 1.06 & 10.65 & 14.83 & 5.46 & 0.33 & 6.84 \\
& OT+\ours & \textbf{27.68} & \textbf{7.84} & \textbf{1.25} & \textbf{12.18} & \textbf{25.67} & \textbf{6.66} & \textbf{0.68} & \textbf{10.64} \\
\midrule
\multirow{4}{*}{ Cyclist@IoU 0.5 } & Direct & 38.46 & 2.04 & 0.00 & 19.82 & 30.16 & 1.56 & 0.00 & 15.65 \\
& Direct+\ours & \textbf{54.41} & \textbf{3.19} & 0.00 & \textbf{28.01} & \textbf{47.57} & \textbf{1.94} & 0.00 & \textbf{24.06} \\
\cmidrule{2-10}
& OT & 43.43 & 2.18 & 0.00 & 22.30 & 14.65 & 1.23 & 0.00 & 8.01 \\
& OT+\ours & \textbf{55.26} & \textbf{3.04} & 0.00 & \textbf{28.75} & \textbf{38.12} & \textbf{1.89} & 0.00 & \textbf{19.52} \\
\bottomrule
\end{tabular}
}
\label{tab:dsvt}
\end{table*} %

\subsection{Additional Qualitative Results}

\begin{figure*}[t]
    \centering
    \includegraphics[width=\linewidth]{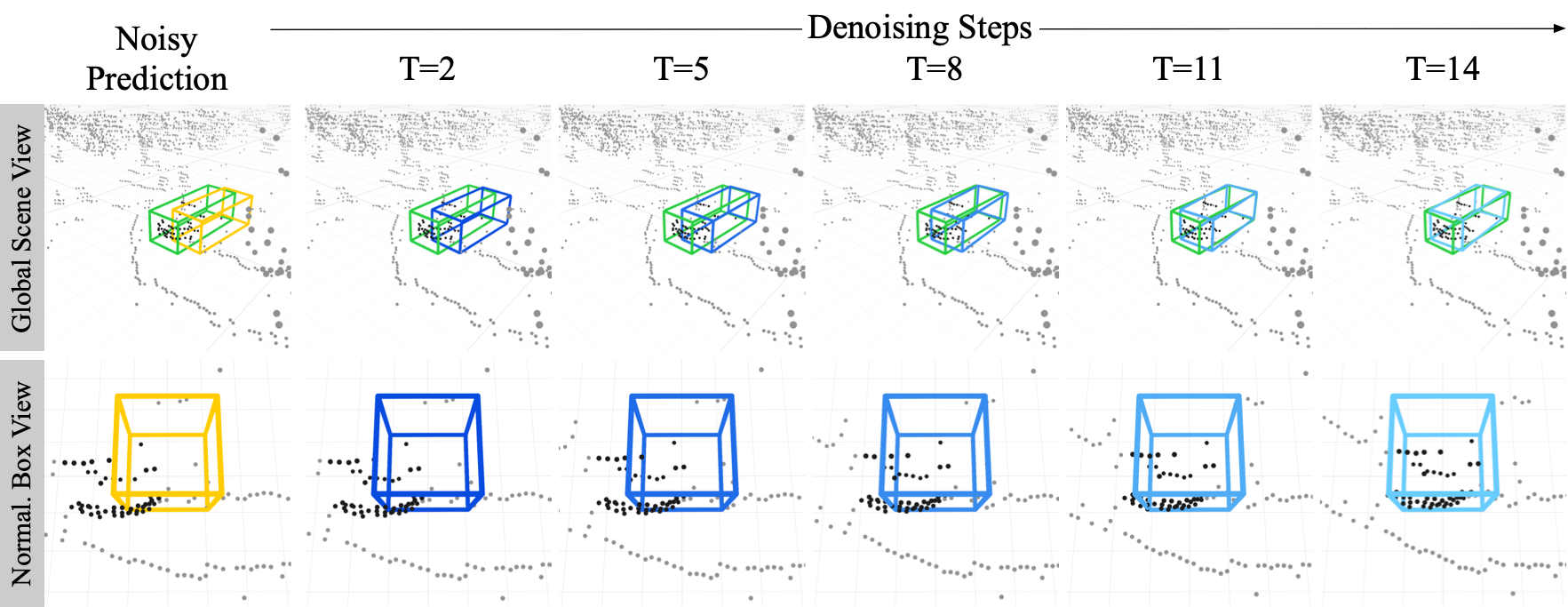}
    \caption{
    \textbf{Box refinement through denoising steps.} We visualize the correction of a noisy prediction, shown in yellow, using \ours, as well as the normalized box view. The detection output is refined iteratively though the denoising steps, resulting in the final, corrected output of our method.
    }
    \label{fig:by-step-normal}
\end{figure*}

We present additional visualizations of box refinement over denoising steps, in both original and normalized views, in \autoref{fig:by-step-normal}. We show the box refinement process of the initial noisy prediction across different denoising steps, up to 14 steps. In addition, we show the points from the normalized box view (NBV) from the perspective of the refined boxes. Observe how the points of the car (colored in black) gradually get corrected to be inside the bounding box.

\begin{figure*}[t]
    \centering
    \includegraphics[width=\linewidth]{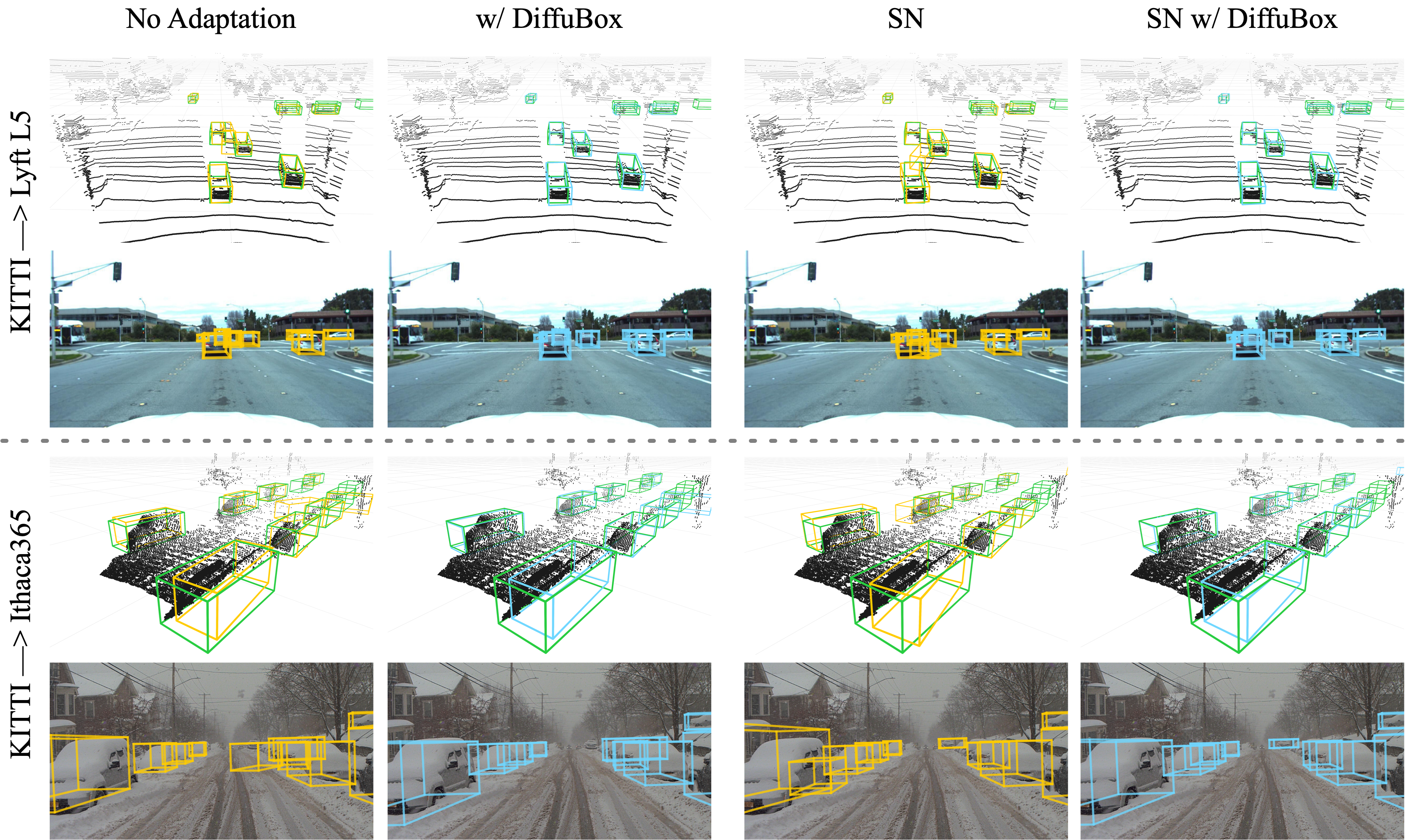}
    \caption{
    \textbf{Additional Qualitative Results of \ours}. We include additional visualizations of 3D object detection on \lyft/\ith before and after \ours's refinement.
    We color the ground truth boxes in green, the detector outputs in yellow, and the boxes refined with \ours in blue. }
    \label{fig:qual-supp}
    \vspace{-1.2\baselineskip}
\end{figure*}

\begin{figure*}[t]
    \centering
    \includegraphics[width=\linewidth]{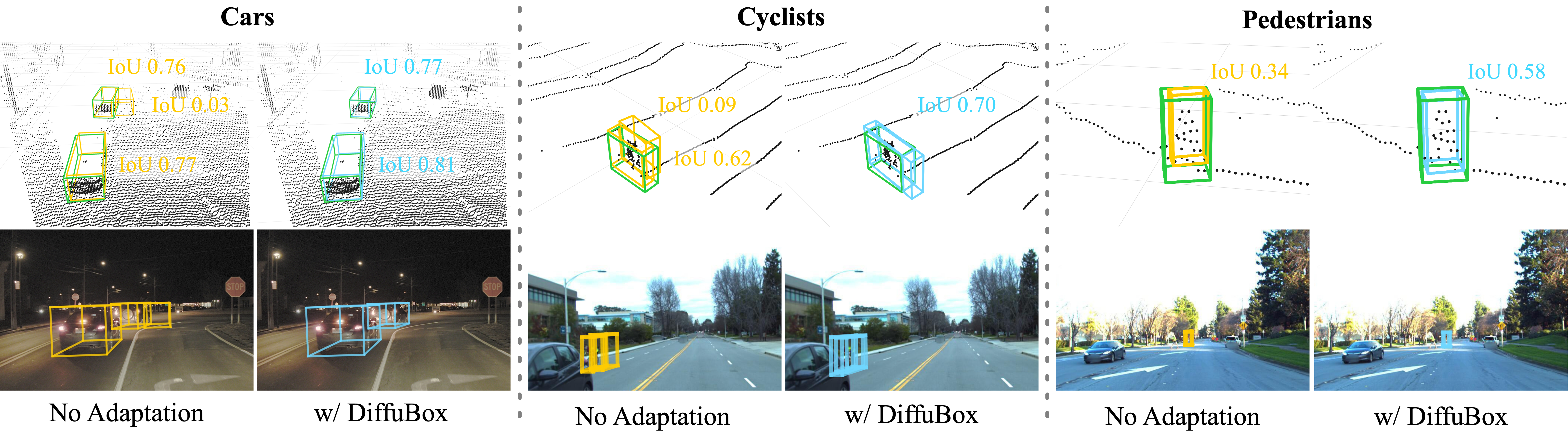}
    \caption{
    \textbf{Additional Qualitative Results of \ours for other classes}. We include additional visualizations of 3D object detection for different classes.  
    We color the ground truth boxes in green, the detector outputs in yellow, and the boxes refined with \ours in blue. }
    \label{fig:qual-supp-multiclass}
    \vspace{-1.2\baselineskip}
\end{figure*}

We present additional visualizations of detections before and after using \ours on another set of scene in \autoref{fig:qual-supp}, and on different classes of traffic participants in \autoref{fig:qual-supp-multiclass}. Observe that \ours's refinement is able to better align the boxes, and the corrected boxes can reduce false positives as they get corrected into the same set of points.

    

\subsection{Improvement Expectations and Upper Bounds}

We validate our claims that improvements in localization is the major source of performance gain in \autoref{tab:perfect-box}. In particular, we show that large gains in performance can be obtained -- across multiple domain adaptation algorithms -- if localization is corrected to the ground truth position. This suggests that there is significant room for improvement in object detection performance from better localization, which \ours{} aims to tackle.
\begin{table}[H]
\centering
\caption{
\textbf{Domain adaptation performance and potential improvement from correcting localization.} 
We report the performance of multiple domain adaptation algorithms (including \emph{Direct}ly applying the model) on a Point-RCNN~\cite{shi2019pointrcnn} detector adapted from KITTI~\cite{kitti_dataset} to Lyft~\cite{lyft_dataset} dataset.
We evaluate mean Average Precision (mAP) on the \emph{Car} category at ${\text{IoU}}_{\text{3D}} = 0.7$.
\emph{Oracle Loc} means assigning ground-truth bounding boxes to any intersecting detections (\ie, infinitesimal ${\text{IoU}}_{\text{3D}}$)\label{tab:perfect-box}
}
\vspace{0.5\baselineskip}
\begin{tabular}{lcccc}
\toprule
Method & \ 0-30m\  & \ 30-50m\  & \ 50-80m\  & \ 0-80m\  \\
\midrule
Direct & 25.76 & 7.84 & 1.04 & 12.07 \\
Direct+Oracle Loc. & \textbf{95.50} & \textbf{80.63} & \textbf{38.24} & \textbf{68.24} \\
Rote-DA~\cite{you2022unsupervised} & 50.63 & 24.80 & 7.07 & 29.19 \\
Rote-DA+Oracle Loc. & \textbf{98.54} & \textbf{80.15} & \textbf{38.55} & \textbf{70.80} \\
SN~\cite{wang2020train} & 70.40 & 32.96 & 6.18 & 36.64 \\
SN+Oracle Loc. & \textbf{96.90} & \textbf{78.19} & \textbf{38.75} & \textbf{70.36} \\
\bottomrule
\end{tabular}
\end{table} %

\subsection{Further Discussion}

\paragraph{Interpretation of the effectiveness of the Statistical Normalization (SN) method.} Recent domain adaptation algorithms typically rely on Statistical Normalization (SN)~\cite{wang2020train} to tackle the domain bias in object size. 
SN's solution is to adapt the source dataset first, rescaling its objects (\ie bounding box dimensions \& points in bounding boxes) to match size statistics in the target domain.
Afterwards, they finetune the object detector on the adapted dataset. 
In short, SN attempts to explicitly construct a new size prior to replace the old one.
Despite great effectiveness, SN is very sensitive to the accuracy of size statistics in the target domain, and requires careful retraining for each new target domain. In contrast, our work is a more general interpretation of size normalization, and instead rely on \textit{shape invariance} to adapt it across all models and domains.

\end{document}